\documentclass[sigconf]{acmart}

\usepackage[linesnumbered,ruled,vlined]{algorithm2e}
\usepackage[T1]{fontenc}
\usepackage{color}
\usepackage{enumitem}
\usepackage{xspace}
\usepackage{multirow}
\newcommand{\ie}{\emph{i.e.,}\xspace}
\newcommand{\eg}{\emph{e.g.,}\xspace}

\setlist[itemize]{itemsep=-0.25ex,leftmargin=2.5ex}
\setlist[enumerate]{itemsep=-0.25ex,leftmargin=2.5ex}

\AtBeginDocument{%
  \providecommand\BibTeX{{%
    \normalfont B\kern-0.5em{\scshape i\kern-0.25em b}\kern-0.8em\TeX}}}

\setcopyright{acmcopyright}
\copyrightyear{2018}
\acmYear{2018}
\acmDOI{10.1145/1122445.1122456}

\acmConference[Woodstock '18]{Woodstock '18: ACM Symposium on Neural
  Gaze Detection}{June 03--05, 2018}{Woodstock, NY}
\acmBooktitle{Woodstock '18: ACM Symposium on Neural Gaze Detection,
  June 03--05, 2018, Woodstock, NY}
\acmPrice{15.00}
\acmISBN{978-1-4503-XXXX-X/18/06}

\begin{document}

\title{Understanding Chinese Video and Language via Contrastive \\ Multimodal Pre-Training }


\author{Chenyi Lei$^{1,2}$, Shixian Luo$^{2}$, Yong Liu$^{3}$, Wanggui He$^{2}$, Jiamang Wang$^{2}$,  \\ Guoxin Wang$^{5,2}$, Haihong Tang$^{2}$, Chunyan Miao$^{4}$, Houqiang Li$^{1}$}
\affiliation{%
  \institution{$^{1}$University of Science and Technology of China, Hefei, Anhui, China\\
  $^{2}$Alibaba Group, Hangzhou, Zhejiang, China \\
  $^{3}$Alibaba-NTU Singapore Joint Research Institute, Nanyang Technological University, Singapore \\
  $^{4}$School of Computer Science and Engineering, Nanyang Technological University, Singapore \\
  $^{5}$Zhejiang University, Hangzhou, Zhejiang, China \\
  }
}
\email{leichy@mail.ustc.edu.cn, huanyi.lsx@alibaba-inc.com, stephenliu@ntu.edu.sg, wanggui.hwg@alibaba-inc.com}
\email{ jiamang.wang@alibaba-inc.com, {xiaogong.wgx,piaoxue}@taobao.com, ascymiao@ntu.edu.sg,lihq@ustc.edu.cn}

\renewcommand{\shortauthors}{PrePrint. Under Review.}

\begin{abstract}

The pre-trained neural models have recently achieved impressive performances in understanding multimodal content. However, it is still very challenging to pre-train neural models for video and language understanding, especially for Chinese video-language data, due to the following reasons. Firstly, existing video-language pre-training algorithms mainly focus on the co-occurrence of words and video frames, but ignore other valuable semantic and structure information of video-language content, \eg sequential order and spatiotemporal relationships. Secondly, there exist conflicts between video sentence alignment and other proxy tasks. Thirdly, there is a lack of large-scale and high-quality Chinese video-language datasets (\eg including 10 million unique videos), which are the fundamental success conditions for pre-training techniques. 


In this work, we propose a novel video-language understanding framework named \textbf{\textsc{Victor}}, which stands for \textbf{VI}deo-language understanding via \textbf{C}ontrastive mul\textbf{T}im\textbf{O}dal p\textbf{R}e-training. Besides general proxy tasks such as masked language modeling, \textsc{Victor} constructs several novel proxy tasks under the contrastive learning paradigm, making the model be more robust and able to capture more complex multimodal semantic and structural relationships from different perspectives. \textsc{Victor} is trained on a large-scale Chinese video-language dataset, including over 10 million complete videos with corresponding high-quality textual descriptions. We apply the pre-trained \textsc{Victor} model to a series of downstream applications and demonstrate its superior performances, comparing against the state-of-the-art pre-training methods such as VideoBERT and UniVL. The codes and trained checkpoints will be publicly available to nourish further developments of the research community.  

\end{abstract}




\maketitle

\section{Introduction}

\begin{figure*}
	\centering
	\includegraphics[width=0.95\textwidth]{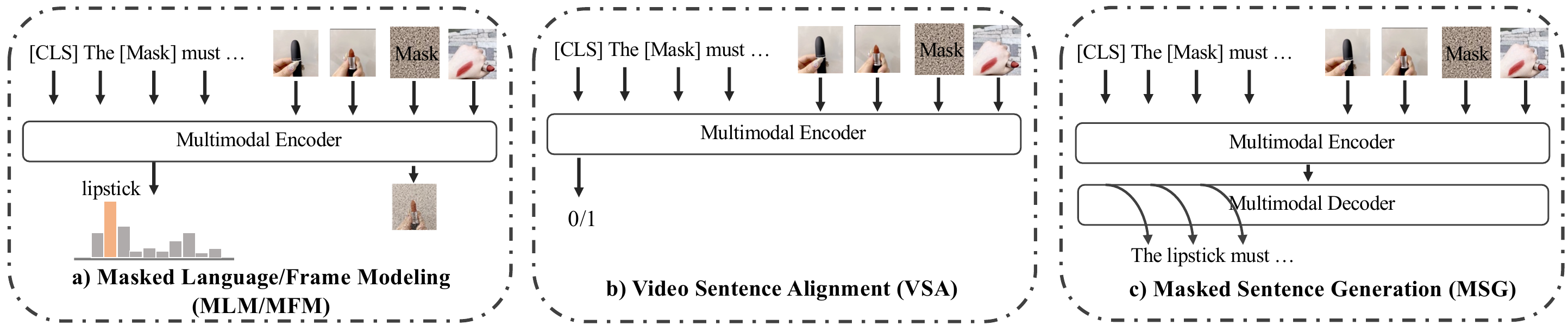}
	\caption{General proxy tasks that are widely used in existing video-language pre-training methods.}
	\label{fig_proxytasks}
\end{figure*}

Inspired by the revolutionary successes of pre-training strategies in natural language processing~\citep{bert,xlnet,t5,ERNIE,roberta}, there has been a steady momentum of breakthroughs in vision-language pre-training techniques~\citep{vlbert,LXMERT,vilbert,uniter,jdbert,clip,dalle,wenlan,pixelbert,oscar,unicoder}. 
However, most of existing vision-language pre-training methods are focusing on image-language pre-training tasks. The pre-training techniques developed for video-language pre-training are still relatively limited~\citep{videobert,univl,ClipBERT,HERO,cbt}. 
VideoBERT~\cite{videobert} and CBT~\cite{cbt} are the pioneering video-language pre-training methods that are developed based on the Transformer~\cite{attallyouneed} structure and three widely used proxy tasks, \ie masked language modeling, masked frame modeling, and video and sentence alignment.   
To support both discriminant and generation tasks, UniVL~\cite{univl} further introduces an encoder-decoder pre-training framework with masked sentence generation task. Figure~\ref{fig_proxytasks} shows an example of these proxy tasks. 



Compared with image-language pre-training, video-language pre-training has the following main challenges. First, videos are dynamic and have more complex spatiotemporal and sequential relationships than static 2D images. 
The spatiotemporal relationship reflects the correlation of video frames, and the sequential relationship verifies the semantic coherence, including visual frames and textual segmentations. Most existing video-language pre-training models are directly extended from image-language pre-training methods without exploiting these unique video and language relationships. 
Second, there are conflicts between the typical implementation of Video Sentence Alignment (VSA)~\cite{videobert} and other proxy tasks. The reason is that mismatched video-text pairs would force the model to fuse irrelevant multimodal signals to understand video and language with other proxy tasks~\cite{vlbert,jdbert}. 
Third, existing video-language datasets are mainly crawled from websites, leading to lots of semantic irrelevant video-text pairs in the dataset. There is a lack of large-scale and high-quality datasets for video-language pre-training, especially Chinese video-language datasets.

To tackle these aforementioned challenges, we firstly propose a novel video-language pre-training framework, namely \textbf{\textsc{Victor}} (\ie \textbf{VI}deo-language understanding via \textbf{C}ontrastive mul\textbf{T}im\textbf{O}dal p\textbf{R}e-training). 
Specifically, \textsc{Victor} is based on the encoder-decoder framework, whose backbone is a stacking of Transformers~\cite{attallyouneed}. The encoder part can be used to learn robust multimodal representations of video and text for downstream discriminant tasks such as classifications and cross-modal retrieval. The decoder part emphasizes the ability of generations, which can be applied to downstream generation tasks such as video caption and summarization. The proxy tasks of \textsc{Victor} fall into two categories: 1) \emph{reconstructive proxy tasks}, and 2) \emph{contrastive proxy tasks}. 
\begin{itemize}
    \item The reconstructive proxy tasks of \textsc{Victor} include two general tasks, \ie Masked Language Modeling (MLM) and Masked Sentence Generation (MSG), which have been used in previous work~\citep{univl}. In addition, we also propose the following two novel reconstructive proxy tasks from the sequential structure perspective, \ie Masked Frame Order Modeling (MFOM) and Masked Sentence Order Modeling (MSOM). The objective is to learn the semantic sequential order information from both video frames and sentence segmentations. 
    
    
    \item The contrastive proxy tasks include three novel tasks, \ie dual Video Sentence Alignment (dual-VSA), intra-Masked Frame Modeling (intra-MFM), and inter-Masked Frame Modeling (inter-MFM). We extend typical VSA~\cite{videobert} to dual-VSA to avoid forcing the model to fuse inaccurate multimodal signals. The intra-MFM and inter-MFM proxy tasks aim to exploit the spatiotemporal supervision signals in the video. Particularly, intra-MFM focuses on the spatial standpoint within a video, and inter-MFM emphasizes the spatiotemporal standpoint among different videos. In this work, we formulate all the contrastive proxy tasks under the contrastive learning paradigm~\citep{moco,simclr}. 
    The basic principle is to make positive/negative query-key pairs similar/dissimilar by the ranking loss or noise contrastive estimation (NCE) loss. Inspired by MoCo~\cite{moco}, we additionally build a dynamic memory queue to track keys across mini-batches for better training.
\end{itemize}
Note that the data pre-processing for MLM, MFOM, and MSOM can be regarded as data augmentations in this work, which makes comparative learning tasks more robust. 

\begin{figure}
	\centering
	\includegraphics[width=0.48\textwidth]{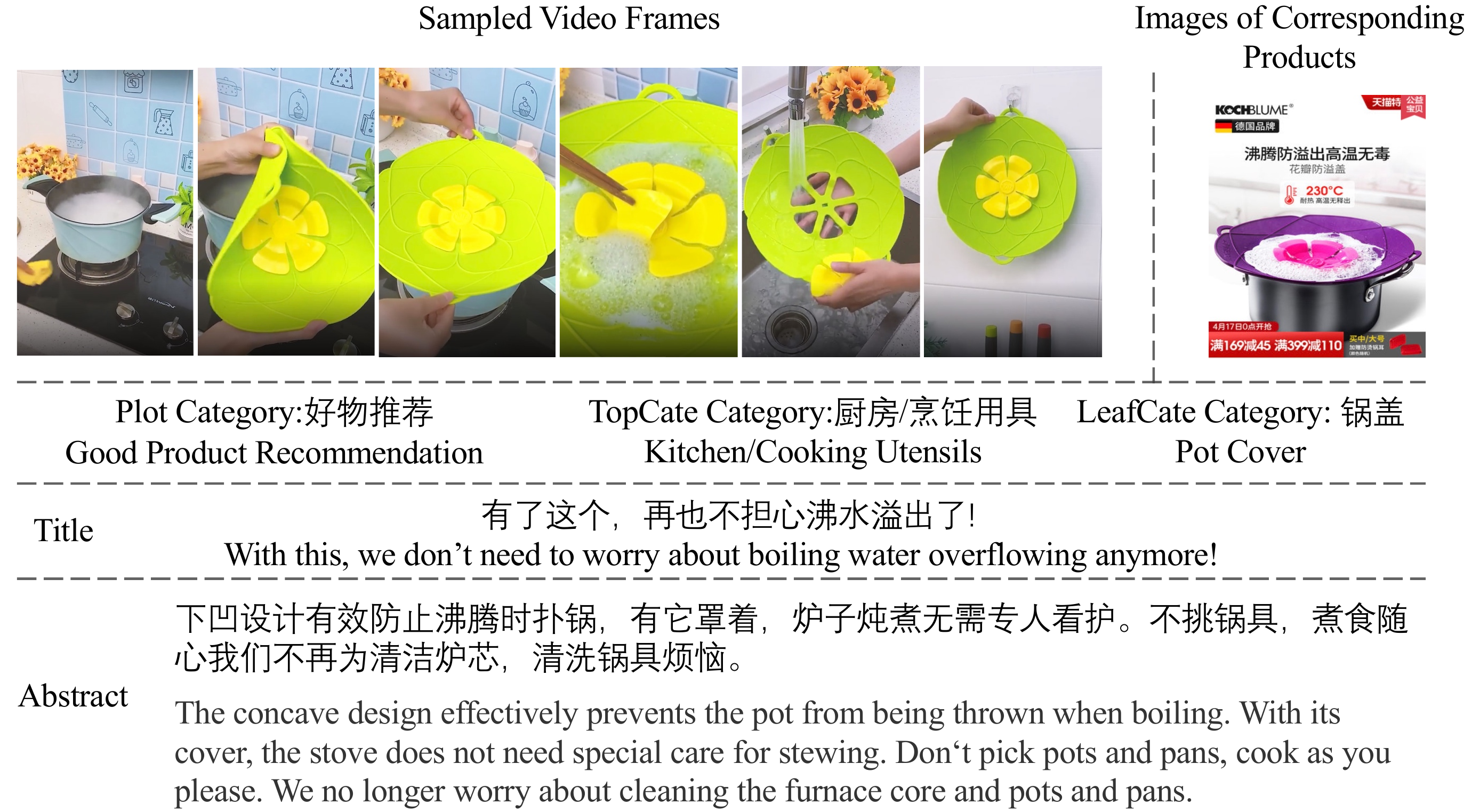}
	\caption{An example of the videos in \textsc{Alivol}-10M dataset. Besides the high-resolution video frames and human-created title, the video also contains a long-text abstract and a related e-commerce product with images. Each video has three types of categories: plot category, coarse-grained product category, and fine-grained product category (respectively denoted by Plot, TopCate, and LeafCate).}
	\vspace{-12pt}
	\label{fig_example}
\end{figure}

To empower the model with rich Chinese knowledge, it is essential to perform pre-training on large-scale and high-quality Chinese video-language datasets. In this work, we also collected \textbf{A} \textbf{L}arge-scale Ch\textbf{I}nese \textbf{V}ide\textbf{O}-\textbf{L}anguage dataset, named \textbf{\textsc{Alivol}-10M}, from one of the world's largest e-commerce platforms. In \textsc{Alivol}-10M, there are more than $10$ million short-videos, instead of video clips cut from complete videos as in existing datasets such as HowTo100M~\cite{howto100}. The total duration of videos in \textsc{Alivol}-10M is about 99 thousand hours. To the best of our knowledge, \textsc{Alivol}-10M is the largest Chinese video-language dataset. Moreover, all the videos in \textsc{Alivol}-10M are carefully produced by professional content creators, and each video is accompanied by a manually written high-quality text description. Furthermore, each video in \textsc{Alivol}-10M also contains various manually annotated information, \eg plot categories, e-commerce categories, and semantic tags.  
These additional knowledge provides several new credible downstream applications for evaluation. Figure~\ref{fig_example} illustrates an example of the videos in \textsc{Alivol}-10M.

The main contributions made in this work are as follows:
\begin{itemize}
 \item We propose a novel encoder-decoder framework, named \textsc{Victor}, for video-language pre-training with novel proxy tasks, including MFOM, MSOM, dual-VSA, intra-MFM, and inter-MFM. \textsc{Victor} is able to capture the sequential structure and spatiotemporal information from the video and language data.
 
 \item We formulate the proposed contrastive proxy tasks, \ie dual-VSA, intra-MFM, and inter-MFM, under the contrastive learning paradigm with an advanced momentum updating strategy.
  
 \item We build to-date the largest Chinese video-language dataset for video-language pre-training,\ie \textsc{Alivol}-10M, which includes over 10 million high-quality e-commerce videos and corresponding human-created text descriptions. 
    
 \item Experimental results on \textsc{Alivol}-10M demonstrate that \textsc{Victor} achieves significant improvements over state-of-the-art (SOTA) video-language pre-training methods, \ie VideoBERT and UniVL, on four different downstream applications. We also conduct an in-depth analysis about the impacts of different proxy tasks on the pre-training performances. 
 The source codes of \textsc{Victor}, pre-trained models, and evaluation dataset will be released to nourish further developments of the research community.
\end{itemize}

\section{Related Work}

%


The large scale pre-training techniques have been widely studied in recent years. In the language domain, Transformer~\cite{attallyouneed} based approaches pre-trained with the masked language modeling (MLM) objective have revolutionized Natural Language Processing (NLP). BERT~\cite{bert}, RoBERTa~\cite{roberta}, ALBERT~\cite{albert}, XLNET~\cite{xlnet}, T5~\cite{t5}, and ERNIE~\cite{ERNIE} are the representative methods which have been used across different NLP tasks. Riding on the successes of BERT-style pre-training models in NLP, numerous visual-linguistic pre-training models have been developed. For example, ViLBERT~\cite{vilbert} and LXMERT~\cite{LXMERT} are the pioneering image-language pre-training methods, which adopt two separate Transformers for encoding image and text independently and a third Transformer for later multimodal fusion. 
Compared with these two-stream frameworks, VL-BERT~\cite{vlbert}, Unicoder-VL~\cite{unicoder}, OSCAR~\cite{oscar}, and UNITER~\cite{uniter} use one shared BERT model to build a single-stream architecture. These models mainly employ MLM and image-text matching as pre-training tasks. Moreover, the recent work DALL-E~\cite{dalle} and TDEN~\cite{jdbert} also make impressive progress on generation tasks, \eg text-to-image generation and image captions. 


For video-language pre-training, VideoBERT~\cite{videobert} and CBT~\cite{cbt} are the pioneers to explore the capability of pre-training models, which are single-stream and two-stream architectures respectively. 
For better pre-training on generation tasks, UniVL~\cite{univl} proposes an encoder-decoder framework based on the two-stream architecture. HERO~\cite{HERO} further encodes multimodal inputs in hierarchical transformers with two new proxy tasks, \ie video-subtitle matching and frame order modeling. ClipBERT~\cite{ClipBERT} proposes to utilize a sparse sampling strategy to model videos from raw pixels. CUPID~\cite{CUPID} is a recent work studying the domain gap between video-language pre-training and fine-tuning. 
These pre-trained models have been applied in various downstream applications, \eg action recognition~\cite{kinect}, video caption~\cite{youcook,autogif}, video retrieval~\cite{msrvtt}, and video question answering~\cite{tgifqa,mq}. The video-language data has more complex semantic structures and relationships than the image-language data, leading video-language pre-training to be a much more challenging research problem. However, existing video-language pre-training works lack detail exploring these unique challenges, such as the spatiotemporal and sequential relationships.

Large-scale and high-quality datasets are the fundamental success condition for large-scale pre-training techniques. For image-language pre-training, there exist numerous large-scale and high-quality datasets with over 100 thousand unique images, including English datasets~\cite{coco,vg} and Chinese datasets~\cite{wenlan,m6} (10 millions level). However, there are only three datasets including over 100 thousand unique videos for video-language pre-training, \ie TGIF~\cite{tgif}, AutoGIF~\cite{autogif}, and HowTo100M~\cite{howto100}. All TGIF, AutoGIF, and HowTo100M are English datasets and crawled from open websites. There exist lots of videos with irrelevant textual descriptions. Moreover, to the best of our knowledge, there are no large-scale Chinese video-language datasets. Thus, there is an urgent need to pre-train, evaluate, and release models on a large-scale and high-quality Chinese video-language dataset.

In contrast to the boom in the image-language pre-training area, pre-training for video and language is still in infancy. Additionally, there is no existing work performing large-scale video-language pre-training on Chinese data. In this work, we aim to propel video-language pre-training in two dimensions: 1) developing a better video-language pre-training model, especially for capturing spatiotemporal and sequential relationships of the video and language data; 2) evaluating and releasing models pre-trained on the large-scale and high-quality Chinese video-language dataset.






\begin{figure*}
	\centering
	\includegraphics[width=0.95\textwidth]{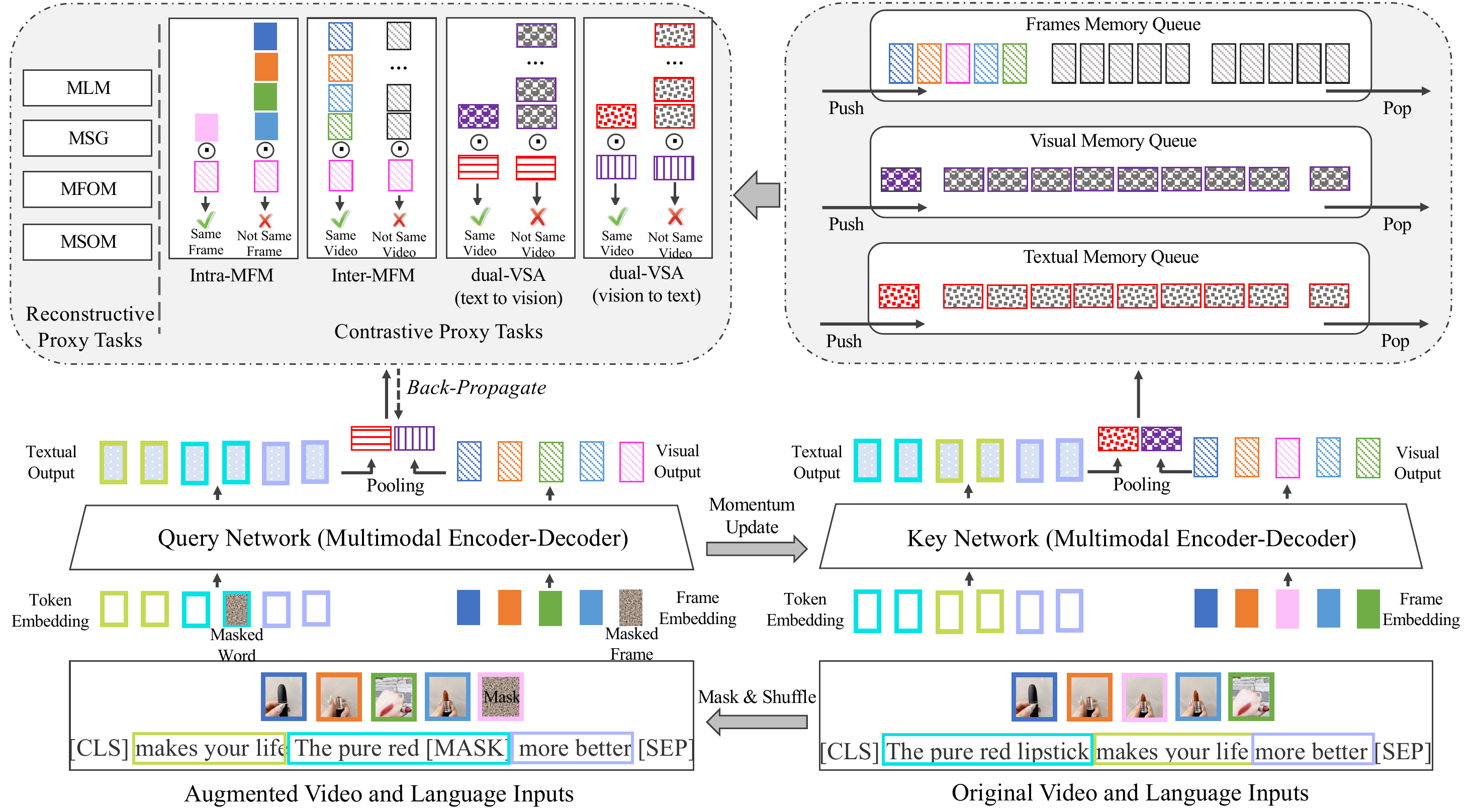}
	\caption{The overall framework of the proposed \textsc{Victor} model. The squares of the same color in the figure represent the associated representations. Please view this figure in color.}
	\label{fig_framework}
\end{figure*}


\section{Method}
\label{sec:method}
In this section, we first introduce an overview of the \textsc{Victor} model and then describe the designed proxy tasks from two perspectives. 
After that, we will introduce the model optimization strategy.

\subsection{Model Overview}


The video-language pre-training problem can be defined as: \textit{given a video and the corresponding textual descriptions as a pair data $\{V,T\}$, pre-training model is to learn joint video and text representation network $f(V,T)$ with the self-supervision approach}. The representation network $f(V,T)$ is expected to maximize the mutual information and preserve the complementary information of multi-modalities. Thus, it can be further fine-tuned for various downstream applications, \eg video classifications and captions. As discussed in recent work in NLP~\cite{t5} and multimodal pre-training~\cite{univl}, the encoder-decoder framework usually achieves the best performances for both discriminant and generation tasks. Motivated by these observations, we also adopt the encoder-decoder architecture as pre-training framework to learn $f(V,T)$ in this work. The detailed architecture of \textsc{Victor} is illustrated as follows.

%
%

\textbf{Input Embedder.} For each textual description $T$, we follow BERT~\cite{bert} and tokenize the input sentence into WordPieces~\cite{wordpiece}, \ie $T=\{w_i | i\in [1,n]\}$, where $n$ is the maximum number of tokens. Two special tokens [CLS] and [SEP] are used to indicate the beginning and ending of the input sentence, respectively. 
The final representation for each token is obtained via summing up its token embedding and position embedding, followed by a layer normalization (LN) layer. For each video $V$, we first sample a frame sequence, \ie $V=\{f_i | i\in [1,m]\}$, where $m$ is the maximum number of frames. Then, we utilize Inception-V4~\cite{inceptionv4} pre-trained on ImageNet~\cite{imagenet} to extract visual features for each frame (denoted as $F_{ori}\in \mathcal{R}^{m\times 1536}$). Next, visual features are fed through a fully connected (FC) layer to be projected into the same lower-dimensional space as token embeddings. 
The final embedding of a frame is then obtained by summing up the outputs of FC layer and the position embeddings, followed by an LN layer. Finally, we have token embeddings $W \in \mathcal{R}^{n\times d}$ for the given text $T$, and frame embeddings $F \in \mathcal{R}^{m\times d}$ for the given video $V$, where $d$ is the hidden size. 

\textbf{Multimodal Transformer Encoder and Decoder.} The multimodal encoder is devised to fully capture the intra-modal relationships and inter-modal interaction. Given the token embedding $W \in \mathcal{R}^{n\times d}$ and frame embeddings $F \in \mathcal{R}^{m\times d}$, we first concatenate them as $[W, F] \in  \mathcal{R}^{(n+m)\times d}$. Then, we further feed $[W, F]$ into a stack of transformer-based encoder blocks~\cite{attallyouneed}, leading to the contextual multimodal representations of each word/frame token $[W_E, F_E] \in  \mathcal{R}^{(n+m)\times d}$. To empower the pre-training model to have generation capability, we also involve a multimodal decoder that learns to auto-regressively reconstruct the input sentence word-by-word conditioned on the input video and masked text with unidirectional attention mechanism~\cite{univl,jdbert}. Here, we implement the multimodal decoder by stacking transformer-based decoder blocks. In particular, each transformer-based decoder block first collects contextual information from all the ``past'' word tokens via unidirectional attention, and then performs cross-attention over all frame tokens for next word prediction.



Figure~\ref{fig_framework} illustrates the overall framework of the \textsc{Victor} model. 
Specifically, \textsc{Victor} consists of a Query Network and a Key Network. Both of them are based on the multimodal encoder-decoder framework mentioned above. The Query Network produces multimodal embeddings with masking and shuffling strategies on both textual and visual inputs, while the Key Network generates multimodal embeddings with the original inputs. After the Query/Key Network, we have multimodal frame embeddings $F_E^{q}$/$F_E^{k}\in \mathcal{R}^{m\times d}$ and multimodal text embeddings $W_E^{q}$/$W_E^{k}\in \mathcal{R}^{n\times d}$. Moreover, we perform max pooling for embeddings to get the general representation for visual and text, \ie $R_{v}\in \mathcal{R}^{d}$ and $R_{t}\in \mathcal{R}^{d}$. Motivated by MoCo~\cite{moco}, the output embeddings of Key Network (\ie $F_E^{k}$, $R_v$, and $R_t$) are pushed into three dynamic memory queues, which continuously generate negative samples for calculating losses for different contrastive proxy tasks. The Query Network is updated by all pre-training losses, and the Key Network is updated conditioned on Query Network weight similar to MoCo. Note all masking and shuffling strategies for Query Network inputs can be regarded as data augmentations for our proposed momentum contrastive pre-training strategy.

\subsection{Reconstructive Proxy Tasks}

\textbf{Masked Language Modeling (MLM).} Motivated by the breakthrough success of the MLM task in NLP, we leverage a multimodal version of MLM for pre-training. Following BERT, we randomly mask $15\%$ text tokens with the special token [MASK] in the sentence, denoted by $W_m$. The goal is to predict these masked tokens based on the video frames $F$ and other tokens $W_{\neg m}$, by minimizing the following negative log-likelihood:
 \begin{equation}
 \label{eq_mlm}
\mathcal{L}_{MLM}(\theta) = -\mathbb{E}_{(W, F) \sim D} \log{P_{\theta}(W_m | W_{\neg m}, F)},
\end{equation}
where $\theta$ denotes the trainable parameters, and each pair $(W, F)$ is sampled from the whole training set $D$.

\textbf{Masked Sentence Order Modeling (MSOF).} The objective of MSOF is to learn the relationships of text tokens from sequential perspective. During the pre-training stage, $15\%$ randomly selected sentences are randomly split into 3 segments, and then all the segments are shuffled by a random permuted order. We let the pre-trained model reorganize these permuted segments, modeled as a $k$-class classification problem, where $k=3!$. Empirically, the sentence reordering task can enable the model to learn the sequential semantic structure of language. Specifically, after the multimodal encoder, the embedding of [CLS] is transformed through an FC layer, followed by a softmax layer, to produce a probability. The final objective is to minimize the following negative log-likelihood:
 \begin{equation}
 \label{eq_msom}
\mathcal{L}_{MSOM}(\theta) = -\mathbb{E}_{(W, F) \sim D} \log{P_{\theta}(y | W, F)},
\end{equation}
where $y$ is the ground truth of shuffled segments.

\textbf{Masked Frame Order Modeling (MFOM).} We also construct a reordering task to learn the sequential semantic structure of the video. For each frame sequence, we randomly select to shuffle $15\%$ frames that are denoted by $F_s=\{f_{s,i}, i\in [1, N_s]\}$, where $N_s$ is the number of shuffled frames. Our objective is to reconstruct the original order of these shuffled frames. The embedding of each shuffled frame is transformed through an FC layer, followed by a softmax layer, to predict the original order index. This is a $m$-class classification problem, where $m$ is the maximum length of frame sequence. We minimize the following negative log-likelihood:
 \begin{equation}
 \label{eq_msom}
\mathcal{L}_{MFOM}(\theta) = -\mathbb{E}_{(W, F) \sim D} \sum^{N_s}_{i=1}\log{P_{\theta}(y_{s,i} | f_{s,i}, W, F)},
\end{equation}
where $y_{s,i}$ is the original order index of the shuffled frame $f_{s,i}$. There exist two main differences between this MFOM proxy task and the frame order modeling task used in the HERO model~\cite{HERO}. One is that HERO only shuffles the unmasked frames. The other is that HERO fuses the frames and corresponding subtitles before frame shuffling. These two differences make the frame order modeling task of HERO much easier. Thus, it is more difficult for HERO to learn the sequential semantic structure of the video.

\textbf{Masked Sentence Generation (MSG).} Inspired by previous work~\cite{t5,univl}, 
we introduce the MSG proxy task to enable the multimodal decoder to auto-regressively reconstruct the input sentence word-by-word. The loss function of the MSG proxy task is defined as follows:
 \begin{equation}
 \label{eq_msg}
\mathcal{L}_{MSG}(\theta) = -\mathbb{E}_{(W, F) \sim D} \log{P_{\theta}(\hat{w}_{i} | \hat{w}_{<i}, W, F)},
\end{equation}
where $W$ is the sentence with $15\%$ masked tokens (c.f. MLM) when pre-training. To endow the model with better generation capacity, we further randomly select $15\%$ sentences with full masked tokens.

\subsection{Contrastive Proxy Tasks}

Besides the input processing of reconstructive proxy tasks, such as masking tokens, we further randomly mask the embeddings of $15\%$ input frames with zeros for the Query Network (Figure~\ref{fig_framework}). We utilize InfoNCE~\cite{infonce} to define the following contrastive loss:
 \begin{equation}
 \label{eq_infonce}
\mathcal{L}_{NCE}(q,k^+,\mathcal{K}^{-}) = -\log{\frac{exp(q^{\top}k^+/\tau)}{exp(q^{\top}k^+/\tau) + \sum_{i=1}^{K}exp(q^{\top}k_i^+/\tau)}},
\end{equation}
where $q$ and $k^+$ are a positive pair for query and key, $\mathcal{K}^{-}$ denotes the negative key set, and $\tau$ is the temperature hyper-parameter. The motivation of such formulation is to train the model to maximize the mutual information of the query and positive key.



\textbf{Intra-Masked Frame Modeling (intra-MFM).} The objective of this proxy task is to guide the model to exploit the inherently spatial variation across frames within a video. In particular, we first feed all encoded embeddings of masked frames from Query Network into an FC layer to project them into vectors with the same dimension as the pre-extracted input frame embeddings (\eg Inception-V4 in our work). We denote the projected embeddings by $\{g_{\theta}(F_{E,i}), i\in [1, N_m]\}$, where $g_{\theta}(\cdot)$ denotes the FC layer, $F_{E,i}$ is the encoded embedding of the i-th masked frame, and $N_m$ is the number of masked frames. Then, our goal is to classify the correct pre-extracted input frame embedding $F_{ori, i}$ comparing to the rest input frame embeddings $\mathcal{K}^{-}_{ori}$ for the given $g_{\theta}(F_{E,i})$. Accordingly, we define the following contrastive loss for this task: 
 \begin{equation}
 \label{eq_intramfm}
\mathcal{L}_{intra-MFM}(\theta) = \mathbb{E}_{(W, F) \sim D} \frac{1}{N_m}\sum_{i=1}^{N_m}  \mathcal{L}_{NCE}(g_{\theta}(F_{E,i}), F_{ori, i}, \mathcal{K}^{-}_{ori}). 
\end{equation}


\textbf{Inter-Masked Frame Modeling (inter-MFM).} In this proxy task, the pre-trained model is learnt to differentiate the augmented frames in the same video or from different videos. From the spatiotemporal perspective, the augmented query frame should be similar to all the other frames in the same video. For each encoded frame $\{F_{E,i}^{q} | i\in [1,m]\}$ from the Query Network, we randomly select one encoded frame $\{F_{E,j}^{k} | j\in [1,m]\}$ of the same video from the Key Network as the positive key. The frame set $\mathcal{K}^{-}_{F}$ from other videos in the dynamic frame memory queue are taken as the negative keys. We define the following contrastive loss to measure the objective of the inter-MFM proxy task: 
 \begin{equation}
 \label{eq_intermfm}
\mathcal{L}_{inter-MFM}(\theta) = \mathbb{E}_{(W, F) \sim D} \frac{1}{m}\sum_{i=1}^{m}  \mathcal{L}_{NCE}(F_{E,i}^{q}, F_{E,j}^{k}, \mathcal{K}^{-}_{F}),  
\end{equation}
where $m$ is the maximum number of sampled frames of a video. 

Such design goes beyond the traditional self-supervision in one video with data augmentation, and fetches more positive frame patches within the same video as supervision for contrastive learning. It sheds new light on objects with temporal evolution (\eg new views/poses of objects). This task elegantly takes the advantage of spatiotemporal structure within videos, and thus strengthens the unsupervised visual and textual feature learning for multimodal video understanding.

\textbf{Dual Video and Sentence Alignment (dual-VSA).} We also involve VSA task to strengthen global representations for video and language understanding. Nevertheless, the typical implementation of VSA (\eg VideoBERT~\cite{bert}), \ie taking the mismatched video and text as inputs and utilizing the output embedding of [CLS] to perform the binary classification, would hurt the performances of downstream discriminant applications. One possible reason is that the introduced mismatched vision-text pairs through shared multimodal encoder would hamper the pre-training of other proxy tasks~\cite{vlbert,jdbert}.
In this work, we only take matched video-text pairs as inputs, and utilize the representation of frames/text to retrieve the representation of corresponding text/frames.

In particular, taking the representation of frames $R_v$ as query, the representation of text $R_t$ from the same video is the positive key, and the text representation set $\mathcal{K}^{-}_{t}$ encoded by the Key Network from other videos in the dynamic textual memory queue are taken as the negative keys. Then, we can define the following loss function:
 \begin{equation}
 \label{eq_dvsa1}
\mathcal{L}_{dual-VSA-v2t}(\theta) = \mathbb{E}_{(W, F) \sim D} \mathcal{L}_{NCE}(R_v, R_t, \mathcal{K}^{-}_{t}).  
\end{equation}
Similarly, we have the dual loss function as follows:
 \begin{equation}
 \label{eq_dvsa2}
\mathcal{L}_{dual-VSA-t2v}(\theta) = \mathbb{E}_{(W, F) \sim D} \mathcal{L}_{NCE}(R_t, R_v, \mathcal{K}^{-}_{v}),  
\end{equation}
where $\mathcal{K}^{-}_{v}$ is the visual representation set encoded by the Key Network from other videos in the dynamic visual memory queue.

\subsection{Model Optimization}
The overall training objective of the proposed pre-training model combines the loss functions of all proxy tasks as follows:
 \begin{equation}
 \begin{aligned}
 \label{eq_loss}
\mathcal{L} =& \mathcal{L}_{MLM} +  \mathcal{L}_{MSOM} +  \mathcal{L}_{MFOM} +  \mathcal{L}_{MSG} +  \\
& \mathcal{L}_{intra-MFM}  +  \mathcal{L}_{inter-MFM} + \\
& \mathcal{L}_{dual-VSA-v2t} + \mathcal{L}_{dual-VSA-t2v}.
\end{aligned}
\end{equation}
In the proposed model, the Query Network is directly optimized by using SGD algorithm to minimize $\mathcal{L}$. The weights of Key Network are accordingly updated conditioned on the Query Network weights via a momentum updating strategy:
 \begin{equation}
 \label{eq_momentum}
\theta_k = \alpha \cdot \theta_k + (1-\alpha) \cdot \theta_q,
\end{equation}
where $\alpha$ stands for momentum value, $\theta_k$ and $\theta_q$ are the parameters of Key Network and Query Network, respectively. This form of contrastive learning can achieve superior performances, due to the large memory queue and the small feature drift during the training process.

\section{Experiments}

In this section, we first introduce the \textsc{Alivol}-10M dataset and experimental settings.
Then, we describe comprehensive experiments on different downstream tasks, including cross-modal video retrieval, video classifications, content-based video recommendation, and multimodal video caption. Finally, extensive ablation studies are presented for in-depth analysis of different pre-training settings.  

\subsection{Pre-Training Dataset}
Most existing datasets for video-language pre-training are in English. To empower the pre-trained model with Chinese knowledge and evaluate the effectiveness of the pre-training mechanism on Chinese data, we build a video-language pre-training dataset, namely \textsc{Alivol}-10M, from one of the world's largest e-commerce platforms. 
Existing video-language datasets are usually collected from open websites~\cite{youcook,msrvtt,tgif,howto100,tgif}, where the data are freely created by users without restriction. This usually leads to the following problems in existing datasets. 
First, there exist many cases where videos are irrelevant to text descriptions. Second, the textual sentences may contain many meaningless tokens, thus may not be fluent in natural language. Third, video and text from open websites may also contain illegal content and user privacy information in some cases. 
Instead, all of the video and text data in \textsc{Alivol}-10M are created by professional creators, \eg merchants, advertising companies, opinion leaders, and professional reviewers, following the standards of the e-commerce platform. 
For example, the videos and texts must be semantically related and must not contain illegal content. As the videos are provided by the e-commerce platform, videos in \textsc{Alivol}-10M mainly describe merchandises in different display forms, \eg merchandise evaluation and situation comedy. This ensures that most videos and texts describe the observed visual content. 
In addition, we also group the videos of \textsc{Alivol}-10M using three different category systems, \ie 13 categories of plot, 153 categories of the coarse-grained product, and 11,529 categories of fine-grained product. We compare some characteristics of \textsc{Alivol}-10M with existing widely used datasets in Table~\ref{tab:dataset_comp}. 
To the best of our knowledge, \textsc{Alivol}-10M is the first large-scale and high-quality dataset for Chinese video-language pre-training. Section A in Appendix displays several typical examples and describes more details of \textsc{Alivol}-10M.

 \begin{table}
  \renewcommand\arraystretch{1}
  \centering
  \caption{Statistics of different video-language datasets.}
  \label{tab:dataset_comp}
  \resizebox{\columnwidth}{!}{
    \begin{tabular}{c|c|c|c|c|c}
      \hline
      Dataset & \# Videos & \# Sentences & Duration & Source & Language \\
      \hline
       YouCook~\cite{youcook} & - & 3K & - & YouTube & English \\
       M-VAD~\cite{mvad} & 92 & 56K & 84h & Movies & English \\
       MPII-MD~\cite{mpii} & 94 & 68K & 41h & Movies & English \\
       MSR-VTT~\cite{msrvtt} & 7,180 & 200K & 40h & Youtube & English \\
       TGIF~\cite{tgif} & 102,068 & 126K & 103h & Tumblr & English \\
       AutoGIF~\cite{autogif} & 163,183 & 165K & - & Web & English \\
       HowTo100M~\cite{howto100} & 1.2M & 136M & 134,472h & Youtube & English \\
       \hline
        \textsc{Alivol}-10M & 10.3M & 11M & 98,801h & E-Commerce & Chinese \\
        \hline
    \end{tabular}
  }
  \flushleft
 \end{table}
 
  \begin{table}
  \renewcommand\arraystretch{1}
  \centering
  \caption{Results of cross-modal video retrieval tasks. R@N denotes the recall of top-N predictions. }
  \label{tab:retrival}
  \resizebox{\columnwidth}{!}{
  \begin{tabular}{c| c c c| c c c}
    \hline
    \multirow{2}{*}{Methods}   & \multicolumn{3}{c|}{Text-based} & \multicolumn{3}{c}{Image-based}  \\
    \cline{2-7}
    &R@1      &R@10   & R@20       & R@1      &R@10   & R@20 \\
    \hline
   VideoBERT & 0.1940 & 0.7132 & 0.8251 & 0.4700 & 0.8373 & 0.8672  \\
   UniVL & 0.1677 & 0.4220 & 0.7221 & 0.4636 & 0.8351 & 0.8901  \\
   \textsc{Victor} & \textbf{0.2425} & \textbf{0.7222} & \textbf{0.8656} & \textbf{0.4770} & \textbf{0.8954} & \textbf{0.9441}  \\
    \hline
    \end{tabular}
  }
  \flushleft
 \end{table}
 
 \subsection{Implementation Details}
We utilize 30,000 token vocabulary and pre-trained text token embeddings of BERT-base-Chinese model\footnote{https://github.com/google-research/bert} to initialize the text token embeddings of \textsc{Victor}. Inception-V4 pre-trained on ImageNet is applied to extract frame features. The maximal input tokens of text are 128, and the maximal number of video frames is 32. For each video, we sample one frame every second. There are 12 and 2 Transformer blocks for multimodal encoder and decoder, respectively. Each Transformer block has 12 heads and 768 hidden size. The size of memory queue in contrastive learning is 65,586. The momentum value $\alpha$ is set to 0.999, and the temperature value $\tau$ is set to 0.7. We adopt Adam~\citep{kingma2014adam} as the optimizer with an initial learning rate $10^{-4}$, and set the batch size to 128 for each GPU. We pre-train the proposed \textsc{Victor} model on 128 NVIDIA Tesla V100 GPUs for 30 epochs, which takes about three days. VideoBERT~\cite{videobert} and UniVL~\cite{univl} are chosen as baseline methods. They are also pre-trained for 30 epochs on \textsc{Alivol}-10M. Moreover, we randomly select 200 thousand videos that are not used in the pre-training stage and use them as the data for downstream tasks. For all downstream tasks, we take 180 thousand videos to fine-tune the pre-trained model and 20 thousand videos to evaluate the performances of different baselines on downstream tasks.

 \subsection{Results on Downstream Tasks}
 
 \subsubsection{Cross-Modal Video Retrieval}
In this work, we focus on two cross-modal video retrieval tasks: text-based and image-based video retrieval. The text-based task aims to retrieve relevant videos from candidates based on the given text query. For the image-based task, we use an image as the given query to retrieve videos. In particular, for \textsc{Alivol}-10M, we utilize the video title and corresponding product image to retrieve relevant videos. For each video, we randomly select 100 videos as the negative ones. Therefore, we obtain 18 million text-video/image-video pairs for fine-tuning and 2 million pairs for evaluating. We introduce the following two fine-tuning strategies for these video retrieval tasks, respectively. 
\begin{itemize}
  \item \textbf{Text-based Retrieval.} The inputs of the text-based video retrieval task are the title and candidate video frames. Similar to the dual-VSA task in pre-training, we also utilize max pooling to obtain the overall representations of text and vision, which are utilized to calculate the matching score based on Cosine distance.
  \item \textbf{Image-based Retrieval.} Besides the query image, the inputs of the image-based video retrieval task include the title and frames of the candidate video. During fine-tuning, we replace the first frame of the candidate video as the query image. The matching score is calculated as the same as in text-based video retrieval.
\end{itemize}

\begin{figure}
	\centering
	\includegraphics[width=0.48\textwidth]{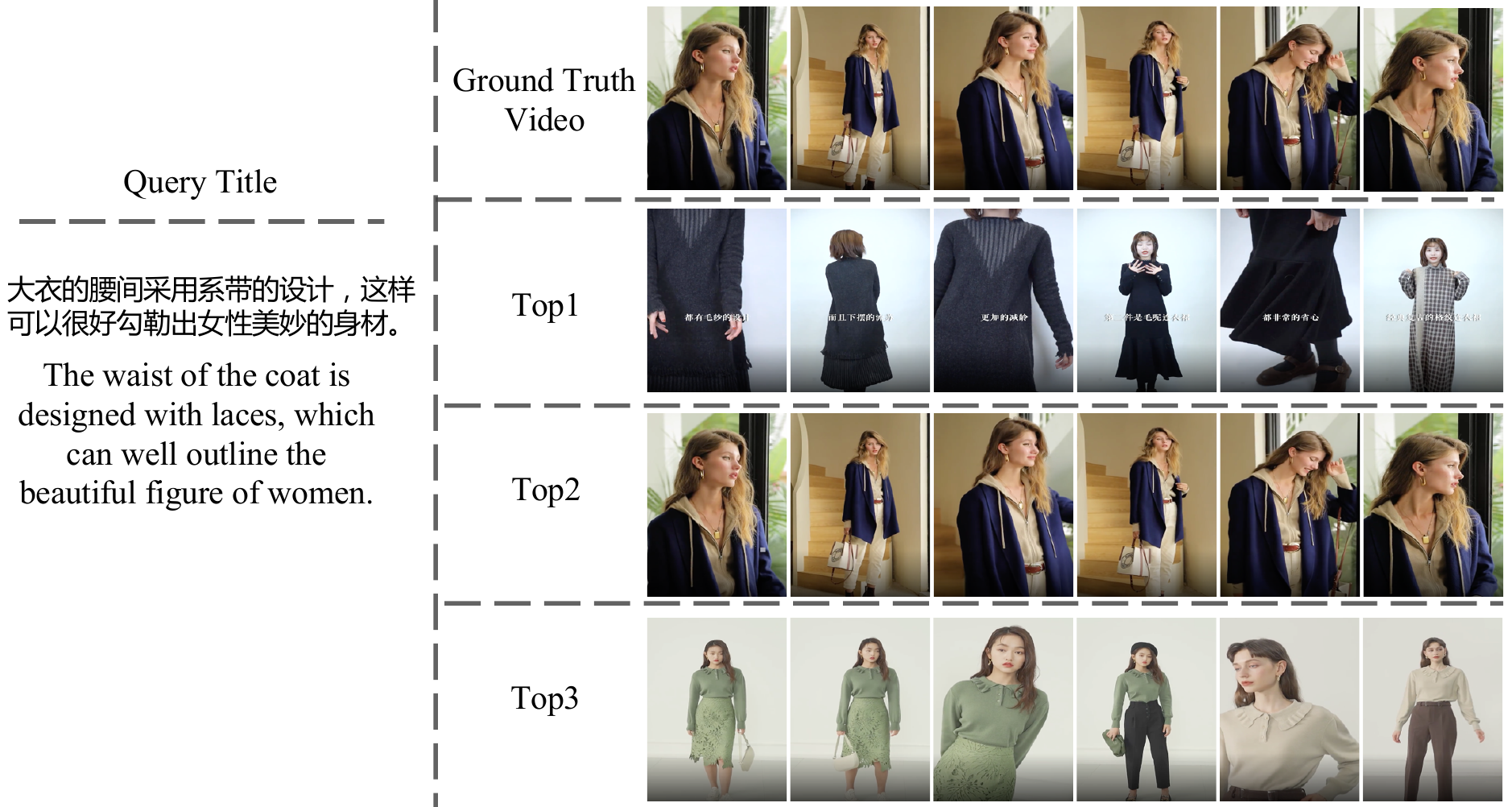}
	\caption{An example result of the text-based video retrieval task, which is fine-tuned based on the pre-trained \textsc{Victor} model.}
	\label{fig_example_t2v}
\end{figure}

\begin{figure}
	\centering
	\includegraphics[width=0.48\textwidth]{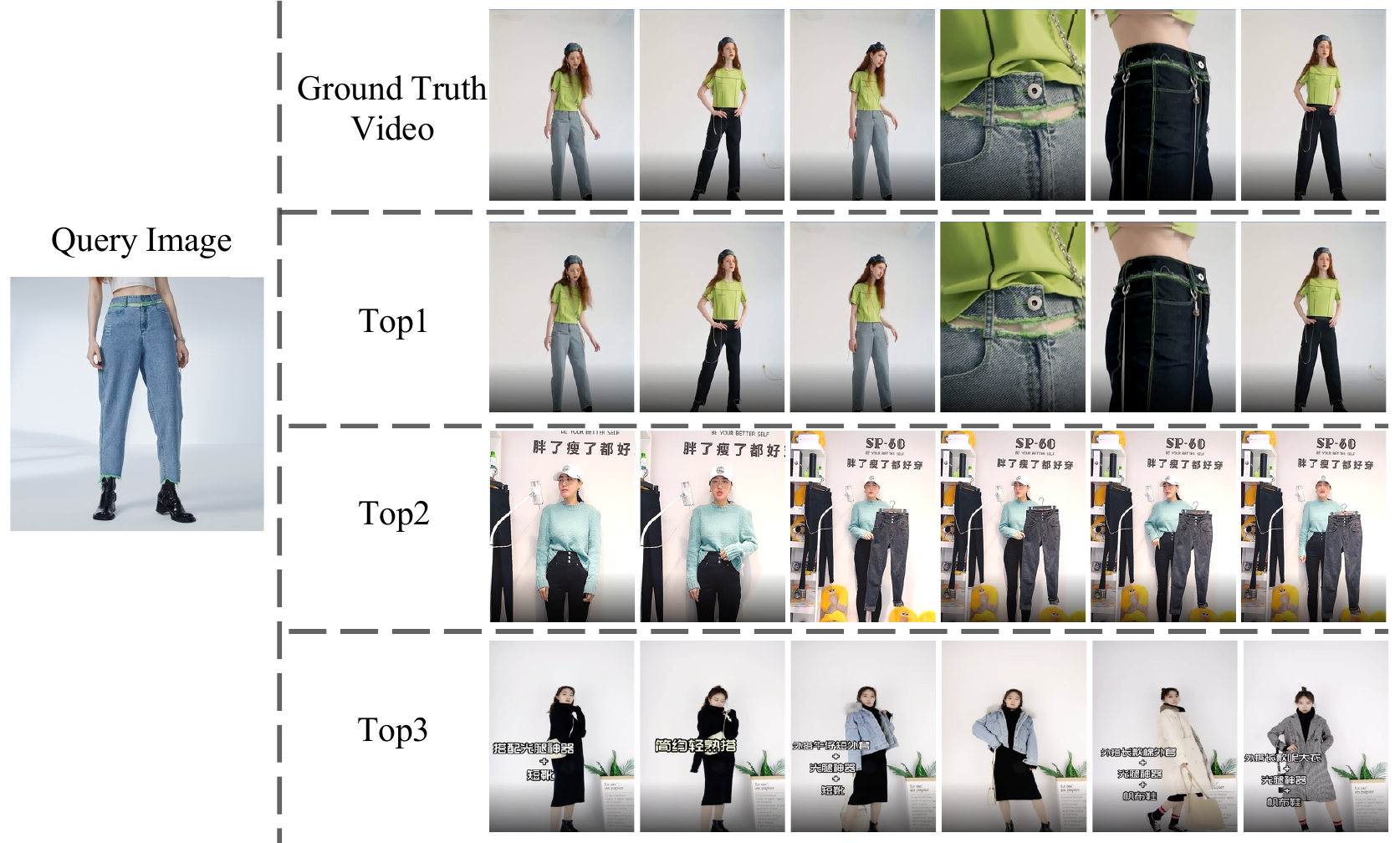}
	\caption{An example result of the image-based video retrieval task, which is fine-tuned based on the pre-trained \textsc{Victor} model.}
	\label{fig_example_i2v}
\end{figure}

From Table~\ref{tab:retrival}, we can note that \textsc{Victor} achieves the best performances comparing to baseline methods with a large extent for two cross-modal video retrieval tasks. In particular, for the image-based video retrieval task, \textsc{Victor} achieves the significant absolute improvement over the second-best method (\ie VideoBERT) by 5.81\%, in terms of R@10. There are two potential reasons. Firstly, this may be caused by the advantage of the dual-VSA proxy task for learning cross-modal semantic representations. Secondly, the inter-MFM task captures the spatiotemporal information of the video to distinguish the query image and candidate frames. Figure~\ref{fig_example_t2v} and Figure~\ref{fig_example_i2v} show case studies for text-based and image-based video retrieval, respectively. Section B in Appendix further introduces more case studies for intuitive understandings.

\subsubsection{Multi-level Video Classification}
For each video in \textsc{Alivol}-10M, it has three different types of category labels from different levels, \ie plot category (Plot), coarse-grained product category (TopCate), and fine-grained product category (LeafCate). 
As there exist great differences in the classification systems of these three types of categories, it is necessary for the pre-trained model to understand videos from different aspects for making an accurate classification. 
In particular, we utilize an additional FC layer after the output embedding of [CLS] to fine-tune pre-trained models for the plot classifier. For the other two multi-level product categories, we add two different FC layers after the output embedding of [CLS] to fine-tune pre-trained models simultaneously. The motivation is that these two multi-level product categories are interrelated. The video classification results are summarized in Table~\ref{tab:dataset_classrec}. For all three category classification tasks, \textsc{Victor} significantly outperforms other two baselines. These results demonstrate the effectiveness of the proposed pre-training strategy for different video classification tasks.

 \begin{table}
  \renewcommand\arraystretch{1}
  \centering
  \caption{Results of three video classification tasks and the content-based video recommendation task. }
  \label{tab:dataset_classrec}
  \resizebox{\columnwidth}{!}{
  \begin{tabular}{c| c c c | c c}
    \hline
    \multirow{3}{*}{Methods}   & \multicolumn{3}{c|}{Classification} & \multicolumn{2}{c}{Recommendation}  \\
    \cline{2-6}
    &Plot      &  TopCate &LeafCate       & -      & -    \\
    &Acc.(\%)      &  Acc.(\%) &Acc.(\%)       & AUC      & HR@3    \\
    \hline
   VideoBERT & 79.46 & 70.14 & 55.43 & 0.7056 & 0.4577 \\
   UniVL & 75.66 & 67.13 & 54.13 & 0.7027 & 0.4523   \\
   \textsc{Victor} & \textbf{81.09} & \textbf{70.50} & \textbf{56.71} & \textbf{0.7104} & \textbf{0.4655}   \\
    \hline
    \end{tabular}
  }
  \flushleft
 \end{table}

  \begin{table}
  \renewcommand\arraystretch{1}
  \centering
  \caption{Results of the multimodal video caption task. }
  \label{tab:dataset_captions}
  \resizebox{\columnwidth}{!}{
    \begin{tabular}{c|c|c|c|c|c}
      \hline
      Methods & BLEU-1 & BLEU-4 & Meteor & Rouge-L & Cider \\
      \hline
       VideoBERT& 23.98 & 9.90 & 11.36 & 20.01 & 0.66 \\
       UniVL & 24.52 & 11.33 & 11.89 & 21.19 & 0.78 \\
       \textsc{Victor} & \textbf{28.38} & \textbf{14.75} & \textbf{13.82} & \textbf{23.89} & \textbf{1.10} \\
        \hline
    \end{tabular}
  }
  \flushleft
 \end{table}
 
    \begin{table*}
  \renewcommand\arraystretch{1}
  \centering
  \caption{Evaluation on proxy tasks and pre-training datasets. All methods are based on our proposed framework. 10M videos in the table denote \textsc{Alivol}-10M, and 100K videos denote a randomly sampled subset of \textsc{Alivol}-10M that contains 100 thousand videos. The best and second-best results are in bold and \underline{underlined}, respectively.}
  \label{tab:dataset_ablation}
  \resizebox{\width}{!}{
  \begin{tabular}{c|l|c| c c | c c c   }
    \hline
   \multirow{3}{*}{Method Number} & \multirow{3}{*}{Proxy Tasks}  & \multirow{3}{*}{\#Pre-Training Videos}& \multicolumn{2}{c|}{Video Retrieval}  & \multicolumn{3}{c}{Video Classification}  \\
    \cline{4-8}
    & & &Text-based      &  Image-Based &Plot       & TopCate     & LeafCate         \\
    & & &R@10               &  R@10              &Acc.(\%)       & Acc.(\%)           & Acc.(\%)              \\
    \hline
    M1 & MLM + MSG                & 10M & 0.6798 & 0.8632 & 78.21 & 70.02 &55.15    \\
    M2 & M1 + MSOM + MFOM & 10M &0.6733 & 0.8689 & 80.42 & 70.04 &56.61  \\
    M3 & M2 + intra/inter-MFM    & 10M   & 0.7015  & 0.8713 & \underline{80.91} & \textbf{70.55} & \textbf{57.15} \\
    M4 & M3 + VSA                    & 10M &  \underline{0.7108} & \underline{0.8764} & 79.80 & 70.2   &55.15   \\
    \hline
    M5& M3 + dual-VSA & 100K & 0.6916  & 0.8641 & 79.17 &  68.08& 55.10  \\
    M6& M3 + dual-VSA   & 10M & \textbf{0.7222} & \textbf{0.8954} & \textbf{81.09} & \underline{70.50} &\underline{56.71}  \\
    \hline
    \end{tabular}
  }
  \flushleft
 \end{table*}
 
 \subsubsection{Content-Based Video Recommendation}
The content-based video recommendation task aims to recommend videos to a user based on the multimodal information of videos viewed by her~\cite{cb:BERT4SessRec, tmm21}. 
Thus, the multimodal representations of videos are crucial for the content-based video recommendation task. We utilize users' video viewing logs on the e-commerce platform to evaluate the performance of this task. Specifically, we collect seven days of user video viewing logs as training data, and the user video viewing logs in the following day are used as test data. To this end, we have about 200 thousand users with 1 million logs for training and about 30 thousand users with 200 thousand logs for evaluation. The DIN~\cite{din} model, which is a well-known recommendation method, is selected as the backbone. The video representations are extracted by pre-trained models, which are the output embeddings of [CLS]. Area Under the receiver operating characteristic Curve (AUC) and Hit Ratio at rank K (denoted by HR@K ) are selected as evaluation metrics. Table~\ref{tab:dataset_classrec} summarizes the video recommendation results. As shown in Table~\ref{tab:dataset_classrec}, \textsc{Victor} achieves the best performances, in terms of both AUC and HR. This again demonstrates the multimodal representing capacity of our pre-trained model. 

\subsubsection{Multimodal Video Caption}
\begin{figure}
	\centering
	\includegraphics[width=0.48\textwidth]{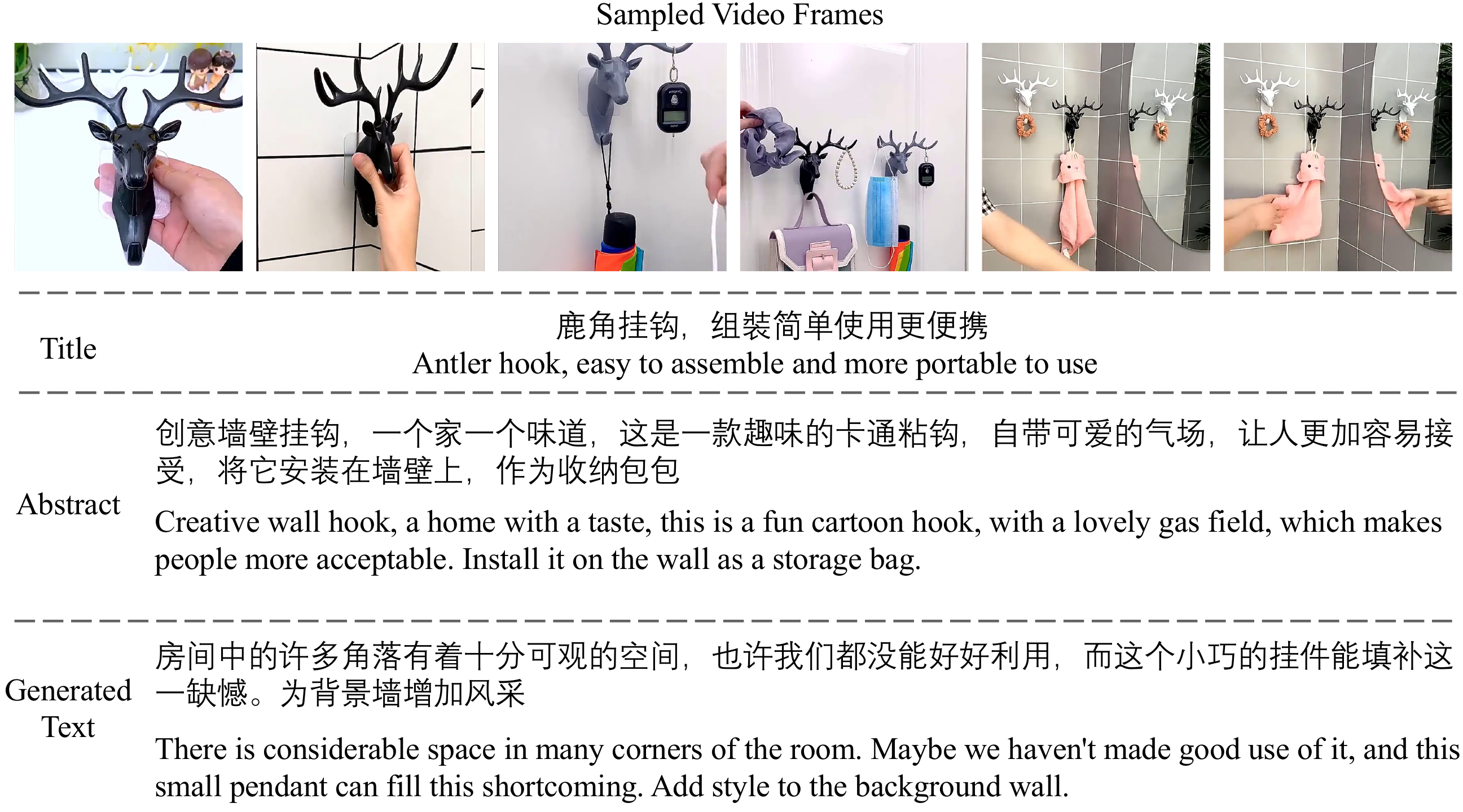}
	\caption{An example result of the video caption task, which is fine-tuned based on the pre-trained \textsc{Victor} model.}
	\label{fig_example_cap}
\end{figure}

The objective of multimodal video caption is to generate a sequence of descriptive sentences based on a given video and some textual hints. In this work, we choose the video title and frames as inputs to generate the video abstract (c.f. Figure~\ref{fig_example}). We use the beam search with setting beam size to 5, and adopt the following corpus-level generation evaluation metrics using the open-source tool\footnote{https://github.com/Maluuba/nlg-eval}, including BLEU-3, BLEU-4, Meteor, Rouge-L, and Cider. Table~\ref{tab:dataset_captions} summarizes the video caption performances achieved by all methods. We can note that our pre-training model \textsc{Victor} outperforms UniVL, which is the SOTA pre-training model for generation task. These results verify the generation capacity of our proposed model. Figure~\ref{fig_example_cap} shows a case study for the multimodal video caption task. More case studies can be found in Section C in Appendix.

 
 \subsection{In-depth Analysis}
We also perform experiments analyzing the effectiveness of our model design, especially focusing on different combinations of the proxy tasks. Table~\ref{tab:dataset_ablation} summarizes the ablation study results on different scale pre-training datasets, under different proxy tasks. Particularly, we use 10M to denote \textsc{Alivol}-10M, and 100K to denote a randomly sampled subset of \textsc{Alivol}-10M that contains 100 thousand videos. From Table~\ref{tab:dataset_ablation}, we can make the following observations. Firstly, M6 consistently outperforms M5. This indicates that the pre-training method benefits a lot from the larger pre-training dataset. It also verifies the contribution of training and releasing the pre-trained model on a large-scale Chinese video-language dataset. Secondly, comparing the performances of M1 and M2, we can note that MSOM and MFOM significantly improve the accuracy of the Plot classification task. This indicates that sequential characteristics of sentence and video frames can benefit the tasks relying on sequential reasoning. Thirdly, the method M3, which considers Intra- and Inter-MFM, not only achieves the best results in TopCate and LeafCate classification, but also achieves significantly better video retrieval accuracy than M1 and M2. This demonstrates the importance of modeling spatiotemporal information of videos. Fourthly, although M4 improves the R@10 of video retrieval tasks, the performances of video classification tasks drop a lot. Instead, the method M6 achieves better results than M4 and comparable results with M3, in video classification tasks. Moreover, M6 achieves the best video retrieval performances. These observations verify the motivation of introducing dual-VSA.

\section{Conclusion and Future Work}
In this paper, we propose a novel encoder-decoder framework, namely \textsc{Victor} (\ie VIdeo-language understanding via Contrastive mulTimOdal pRe-training), for video-language pre-training with novel reconstructive proxy tasks and contrastive proxy tasks. To support Chinese video-language pre-training,  we also build to-date the largest Chinese video-language dataset \textsc{Alivol}-10M, which includes more than 10 million videos and corresponding text descriptions. Experimental results on \textsc{Alivol}-10M demonstrate that the proposed \textsc{Victor} model effectively exploits the multimodal information of the video-language data to achieve better performances than STOA video-language pre-training methods in downstream tasks, including cross-modal video retrieval, multi-level video classification, content-based video recommendation, and multimodal video caption. 
For future work, we would like to pre-train the proposed \textsc{Victor} on large-scale English datasets, \eg HowTo100M~\cite{howto100}, and study its performances in different downstream tasks.

\bibliographystyle{ACM-Reference-Format}
\bibliography{sample-base}


\begin{thebibliography}{47}


\ifx \showCODEN    \undefined \def \showCODEN     #1{\unskip}     \fi
\ifx \showDOI      \undefined \def \showDOI       #1{#1}\fi
\ifx \showISBNx    \undefined \def \showISBNx     #1{\unskip}     \fi
\ifx \showISBNxiii \undefined \def \showISBNxiii  #1{\unskip}     \fi
\ifx \showISSN     \undefined \def \showISSN      #1{\unskip}     \fi
\ifx \showLCCN     \undefined \def \showLCCN      #1{\unskip}     \fi
\ifx \shownote     \undefined \def \shownote      #1{#1}          \fi
\ifx \showarticletitle \undefined \def \showarticletitle #1{#1}   \fi
\ifx \showURL      \undefined \def \showURL       {\relax}        \fi
\providecommand\bibfield[2]{#2}
\providecommand\bibinfo[2]{#2}
\providecommand\natexlab[1]{#1}
\providecommand\showeprint[2][]{arXiv:#2}

\bibitem[\protect\citeauthoryear{Atousa~Torabi}{Atousa~Torabi}{2015}]%
        {mvad}
\bibfield{author}{\bibinfo{person}{Hugo Larochelle Aaron~Courville
  Atousa~Torabi, Christopher~Pal}.} \bibinfo{year}{2015}\natexlab{}.
\newblock \bibinfo{title}{Using Descriptive Video Services to Create a Large
  Data Source for Video Annotation Research}.
\newblock
\newblock
\showeprint[arxiv]{1503.01070}


\bibitem[\protect\citeauthoryear{Chen, Kornblith, Norouzi, and Hinton}{Chen
  et~al\mbox{.}}{2020a}]%
        {simclr}
\bibfield{author}{\bibinfo{person}{Ting Chen}, \bibinfo{person}{Simon
  Kornblith}, \bibinfo{person}{Mohammad Norouzi}, {and}
  \bibinfo{person}{Geoffrey Hinton}.} \bibinfo{year}{2020}\natexlab{a}.
\newblock \showarticletitle{A Simple Framework for Contrastive Learning of
  Visual Representations}. In \bibinfo{booktitle}{\emph{ICML}}. IEEE.
\newblock


\bibitem[\protect\citeauthoryear{Chen, Lei, Liu, Wang, Tang, Zha, and Li}{Chen
  et~al\mbox{.}}{2021}]%
        {tmm21}
\bibfield{author}{\bibinfo{person}{Xusong Chen}, \bibinfo{person}{Chenyi Lei},
  \bibinfo{person}{Dong Liu}, \bibinfo{person}{Guoxin Wang},
  \bibinfo{person}{Haihong Tang}, \bibinfo{person}{Zheng-Jun Zha}, {and}
  \bibinfo{person}{Houqiang Li}.} \bibinfo{year}{2021}\natexlab{}.
\newblock \showarticletitle{E-Commerce Storytelling Recommendation Using
  Attentional Domain-Transfer Network and Adversarial Pre-Training}. In
  \bibinfo{booktitle}{\emph{Transactions on Multimedia}}.
  \bibinfo{publisher}{IEEE}.
\newblock


\bibitem[\protect\citeauthoryear{Chen, Liu, Lei, Li, Zha, and Xiong}{Chen
  et~al\mbox{.}}{2019}]%
        {cb:BERT4SessRec}
\bibfield{author}{\bibinfo{person}{Xusong Chen}, \bibinfo{person}{Dong Liu},
  \bibinfo{person}{Chenyi Lei}, \bibinfo{person}{Rui Li},
  \bibinfo{person}{Zheng-Jun Zha}, {and} \bibinfo{person}{Zhiwei Xiong}.}
  \bibinfo{year}{2019}\natexlab{}.
\newblock \showarticletitle{BERT4SessRec: Content-Based Video Relevance
  Prediction with Bidirectional Encoder Representations from Transformer}. In
  \bibinfo{booktitle}{\emph{MM}}. ACM, \bibinfo{pages}{2597--2601}.
\newblock


\bibitem[\protect\citeauthoryear{Chen, Li, Yu, Kholy, Ahmed, Gan, Cheng, and
  Liu}{Chen et~al\mbox{.}}{2020b}]%
        {uniter}
\bibfield{author}{\bibinfo{person}{Yen-Chun Chen}, \bibinfo{person}{Linjie Li},
  \bibinfo{person}{Licheng Yu}, \bibinfo{person}{Ahmed~El Kholy},
  \bibinfo{person}{Faisal Ahmed}, \bibinfo{person}{Zhe Gan},
  \bibinfo{person}{Yu Cheng}, {and} \bibinfo{person}{Jingjing Liu}.}
  \bibinfo{year}{2020}\natexlab{b}.
\newblock \showarticletitle{UNITER: UNiversal Image-TExt Representation
  Learning}. In \bibinfo{booktitle}{\emph{ECCV}}.
\newblock


\bibitem[\protect\citeauthoryear{Das, Xu, Doell, and Corso}{Das
  et~al\mbox{.}}{2013}]%
        {youcook}
\bibfield{author}{\bibinfo{person}{Pradipto Das}, \bibinfo{person}{Chenliang
  Xu}, \bibinfo{person}{Richard~F. Doell}, {and} \bibinfo{person}{Jason~J.
  Corso}.} \bibinfo{year}{2013}\natexlab{}.
\newblock \showarticletitle{A Thousand Frames in Just a Few Words: Lingual
  Description of Videos through Latent Topics and Sparse Object Stitching}. In
  \bibinfo{booktitle}{\emph{CVPR}}. IEEE.
\newblock


\bibitem[\protect\citeauthoryear{Deng, Dong, Socher, Li, Li, and Fei-Fei}{Deng
  et~al\mbox{.}}{2009}]%
        {imagenet}
\bibfield{author}{\bibinfo{person}{Jia Deng}, \bibinfo{person}{Wei Dong},
  \bibinfo{person}{Richard Socher}, \bibinfo{person}{Li-Jia Li},
  \bibinfo{person}{Kai Li}, {and} \bibinfo{person}{Li Fei-Fei}.}
  \bibinfo{year}{2009}\natexlab{}.
\newblock \showarticletitle{ImageNet: A large-scale hierarchical image
  database}. In \bibinfo{booktitle}{\emph{CVPR}}. IEEE.
\newblock


\bibitem[\protect\citeauthoryear{Devlin, Chang, Lee, and Toutanova}{Devlin
  et~al\mbox{.}}{2018}]%
        {bert}
\bibfield{author}{\bibinfo{person}{Jacob Devlin}, \bibinfo{person}{Ming-Wei
  Chang}, \bibinfo{person}{Kenton Lee}, {and} \bibinfo{person}{Kristina
  Toutanova}.} \bibinfo{year}{2018}\natexlab{}.
\newblock \bibinfo{title}{BERT: Pre-training of Deep Bidirectional Transformers
  for Language Understanding}.
\newblock
\newblock
\showeprint[arxiv]{1810.04805}


\bibitem[\protect\citeauthoryear{Huang, Zeng, Liu, Fu, and Fu}{Huang
  et~al\mbox{.}}{2020}]%
        {pixelbert}
\bibfield{author}{\bibinfo{person}{Zhicheng Huang}, \bibinfo{person}{Zhaoyang
  Zeng}, \bibinfo{person}{Bei Liu}, \bibinfo{person}{Dongmei Fu}, {and}
  \bibinfo{person}{Jianlong Fu}.} \bibinfo{year}{2020}\natexlab{}.
\newblock \bibinfo{title}{Pixel-BERT: Aligning Image Pixels with Text by Deep
  Multi-Modal Transformers}.
\newblock
\newblock
\showeprint[arxiv]{2004.00849}


\bibitem[\protect\citeauthoryear{Huo, Zhang, Liu, Lu, Gao, Yang, Wen, Zhang,
  Xu, Zheng, et~al\mbox{.}}{Huo et~al\mbox{.}}{2021}]%
        {wenlan}
\bibfield{author}{\bibinfo{person}{Yuqi Huo}, \bibinfo{person}{Manli Zhang},
  \bibinfo{person}{Guangzhen Liu}, \bibinfo{person}{Haoyu Lu},
  \bibinfo{person}{Yizhao Gao}, \bibinfo{person}{Guoxing Yang},
  \bibinfo{person}{Jingyuan Wen}, \bibinfo{person}{Heng Zhang},
  \bibinfo{person}{Baogui Xu}, \bibinfo{person}{Weihao Zheng}, {et~al\mbox{.}}}
  \bibinfo{year}{2021}\natexlab{}.
\newblock \showarticletitle{WenLan: Bridging vision and language by large-scale
  multi-modal pre-training}.
\newblock \bibinfo{journal}{\emph{arXiv preprint arXiv:2103.06561}}
  (\bibinfo{year}{2021}).
\newblock


\bibitem[\protect\citeauthoryear{Jang, Song, Kim, Yu, Kim, and Kim}{Jang
  et~al\mbox{.}}{2019}]%
        {tgifqa}
\bibfield{author}{\bibinfo{person}{Yunseok Jang}, \bibinfo{person}{Yale Song},
  \bibinfo{person}{Chris~Dongjoo Kim}, \bibinfo{person}{Youngjae Yu},
  \bibinfo{person}{Youngjin Kim}, {and} \bibinfo{person}{Gunhee Kim}.}
  \bibinfo{year}{2019}\natexlab{}.
\newblock \showarticletitle{Video Question Answering with Spatio-Temporal
  Reasoning}. In \bibinfo{booktitle}{\emph{IJCV}}.
\newblock


\bibitem[\protect\citeauthoryear{Kaiming~He}{Kaiming~He}{2020}]%
        {moco}
\bibfield{author}{\bibinfo{person}{Yuxin Wu Saining Xie Ross~Girshick
  Kaiming~He, Haoqi~Fan}.} \bibinfo{year}{2020}\natexlab{}.
\newblock \showarticletitle{Momentum Contrast for Unsupervised Visual
  Representation Learning}. In \bibinfo{booktitle}{\emph{CVPR}}. IEEE.
\newblock


\bibitem[\protect\citeauthoryear{Kay, Carreira, Simonyan, Zhang, Hillier,
  Vijayanarasimhan, Viola, Green, Back, Natsev, Suleyman, and Zisserman}{Kay
  et~al\mbox{.}}{2017}]%
        {kinect}
\bibfield{author}{\bibinfo{person}{Will Kay}, \bibinfo{person}{Joao Carreira},
  \bibinfo{person}{Karen Simonyan}, \bibinfo{person}{Brian Zhang},
  \bibinfo{person}{Chloe Hillier}, \bibinfo{person}{Sudheendra
  Vijayanarasimhan}, \bibinfo{person}{Fabio Viola}, \bibinfo{person}{Tim
  Green}, \bibinfo{person}{Trevor Back}, \bibinfo{person}{Paul Natsev},
  \bibinfo{person}{Mustafa Suleyman}, {and} \bibinfo{person}{Andrew
  Zisserman}.} \bibinfo{year}{2017}\natexlab{}.
\newblock \bibinfo{title}{The Kinetics Human Action Video Dataset}.
\newblock
\newblock
\showeprint[arxiv]{1705.06950}


\bibitem[\protect\citeauthoryear{Kingma and Ba}{Kingma and Ba}{2014}]%
        {kingma2014adam}
\bibfield{author}{\bibinfo{person}{Diederik~P Kingma} {and}
  \bibinfo{person}{Jimmy Ba}.} \bibinfo{year}{2014}\natexlab{}.
\newblock \showarticletitle{Adam: A method for stochastic optimization}.
\newblock \bibinfo{journal}{\emph{arXiv preprint arXiv:1412.6980}}
  (\bibinfo{year}{2014}).
\newblock


\bibitem[\protect\citeauthoryear{Krishna, Zhu, Groth, Johnson, Hata, Kravitz,
  Chen, Kalantidis, Li, Shamma, Bernstein, and Li}{Krishna
  et~al\mbox{.}}{2016}]%
        {vg}
\bibfield{author}{\bibinfo{person}{Ranjay Krishna}, \bibinfo{person}{Yuke Zhu},
  \bibinfo{person}{Oliver Groth}, \bibinfo{person}{Justin Johnson},
  \bibinfo{person}{Kenji Hata}, \bibinfo{person}{Joshua Kravitz},
  \bibinfo{person}{Stephanie Chen}, \bibinfo{person}{Yannis Kalantidis},
  \bibinfo{person}{Li-Jia Li}, \bibinfo{person}{David~A. Shamma},
  \bibinfo{person}{Michael~S. Bernstein}, {and} \bibinfo{person}{Fei-Fei Li}.}
  \bibinfo{year}{2016}\natexlab{}.
\newblock \bibinfo{title}{Visual Genome: Connecting Language and Vision Using
  Crowdsourced Dense Image Annotations}.
\newblock
\newblock
\showeprint[arxiv]{1602.07332}


\bibitem[\protect\citeauthoryear{Lei, Wu, Liu, Li, Wang, Tang, and Li}{Lei
  et~al\mbox{.}}{2020}]%
        {mq}
\bibfield{author}{\bibinfo{person}{Chenyi Lei}, \bibinfo{person}{Lei Wu},
  \bibinfo{person}{Dong Liu}, \bibinfo{person}{Zhao Li},
  \bibinfo{person}{Guoxin Wang}, \bibinfo{person}{Haihong Tang}, {and}
  \bibinfo{person}{Houqiang Li}.} \bibinfo{year}{2020}\natexlab{}.
\newblock \showarticletitle{Multi-Question Learning for Visual Question
  Answering}. In \bibinfo{booktitle}{\emph{AAAI}}.
\newblock


\bibitem[\protect\citeauthoryear{Lei, Li, Zhou, Gan, Berg, Bansal, and Liu}{Lei
  et~al\mbox{.}}{2021}]%
        {ClipBERT}
\bibfield{author}{\bibinfo{person}{Jie Lei}, \bibinfo{person}{Linjie Li},
  \bibinfo{person}{Luowei Zhou}, \bibinfo{person}{Zhe Gan},
  \bibinfo{person}{Tamara~L. Berg}, \bibinfo{person}{Mohit Bansal}, {and}
  \bibinfo{person}{Jingjing Liu}.} \bibinfo{year}{2021}\natexlab{}.
\newblock \showarticletitle{Less is More: ClipBERT for Video-and-Language
  Learning via Sparse Sampling}. In \bibinfo{booktitle}{\emph{CVPR}}. IEEE.
\newblock


\bibitem[\protect\citeauthoryear{Li, Duan, Fang, Gong, Jiang, and Zhou}{Li
  et~al\mbox{.}}{2020b}]%
        {unicoder}
\bibfield{author}{\bibinfo{person}{Gen Li}, \bibinfo{person}{Nan Duan},
  \bibinfo{person}{Yuejian Fang}, \bibinfo{person}{Ming Gong},
  \bibinfo{person}{Daxin Jiang}, {and} \bibinfo{person}{Ming Zhou}.}
  \bibinfo{year}{2020}\natexlab{b}.
\newblock \showarticletitle{Unicoder-VL: A Universal Encoder for Vision and
  Language by Cross-modal Pre-training}. In \bibinfo{booktitle}{\emph{AAAI}}.
\newblock


\bibitem[\protect\citeauthoryear{Li, Chen, Cheng, Gan, Yu, and Liu}{Li
  et~al\mbox{.}}{2020a}]%
        {HERO}
\bibfield{author}{\bibinfo{person}{Linjie Li}, \bibinfo{person}{Yen-Chun Chen},
  \bibinfo{person}{Yu Cheng}, \bibinfo{person}{Zhe Gan},
  \bibinfo{person}{Licheng Yu}, {and} \bibinfo{person}{Jingjing Liu}.}
  \bibinfo{year}{2020}\natexlab{a}.
\newblock \showarticletitle{HERO: Hierarchical Encoder for Video+Language
  Omni-representation Pre-training}. In \bibinfo{booktitle}{\emph{EMNLP}}.
\newblock


\bibitem[\protect\citeauthoryear{Li, Yin, Li, Zhang, Hu, Zhang, Wang, Hu, Dong,
  Wei, Choi, and Gao}{Li et~al\mbox{.}}{2020c}]%
        {oscar}
\bibfield{author}{\bibinfo{person}{Xiujun Li}, \bibinfo{person}{Xi Yin},
  \bibinfo{person}{Chunyuan Li}, \bibinfo{person}{Pengchuan Zhang},
  \bibinfo{person}{Xiaowei Hu}, \bibinfo{person}{Lei Zhang},
  \bibinfo{person}{Lijuan Wang}, \bibinfo{person}{Houdong Hu},
  \bibinfo{person}{Li Dong}, \bibinfo{person}{Furu Wei}, \bibinfo{person}{Yejin
  Choi}, {and} \bibinfo{person}{Jianfeng Gao}.}
  \bibinfo{year}{2020}\natexlab{c}.
\newblock \showarticletitle{Oscar: Object-Semantics Aligned Pre-training for
  Vision-Language Tasks}. In \bibinfo{booktitle}{\emph{ECCV}}.
\newblock


\bibitem[\protect\citeauthoryear{Li, Pan, Yao, Chen, and Mei}{Li
  et~al\mbox{.}}{2021}]%
        {jdbert}
\bibfield{author}{\bibinfo{person}{Yehao Li}, \bibinfo{person}{Yingwei Pan},
  \bibinfo{person}{Ting Yao}, \bibinfo{person}{Jingwen Chen}, {and}
  \bibinfo{person}{Tao Mei}.} \bibinfo{year}{2021}\natexlab{}.
\newblock \showarticletitle{Scheduled Sampling in Vision-Language Pretraining
  with Decoupled Encoder-Decoder Network}. In \bibinfo{booktitle}{\emph{AAAI}}.
\newblock


\bibitem[\protect\citeauthoryear{Li, Song, Cao, Tetreault, Goldberg, Jaimes,
  and Luo}{Li et~al\mbox{.}}{2016}]%
        {tgif}
\bibfield{author}{\bibinfo{person}{Yuncheng Li}, \bibinfo{person}{Yale Song},
  \bibinfo{person}{Liangliang Cao}, \bibinfo{person}{Joel Tetreault},
  \bibinfo{person}{Larry Goldberg}, \bibinfo{person}{Alejandro Jaimes}, {and}
  \bibinfo{person}{Jiebo Luo}.} \bibinfo{year}{2016}\natexlab{}.
\newblock \bibinfo{title}{TGIF: A New Dataset and Benchmark on Animated GIF
  Description}.
\newblock
\newblock
\showeprint[arxiv]{1604.02748}


\bibitem[\protect\citeauthoryear{Lin, Men, Yang, Zhou, Ding, Zhang, Wang, Wang,
  Jiang, Jia, Zhang, Zhang, Zou, Li, Deng, Liu, Xue, Zhou, Ma, Yu, Li, Lin,
  Zhou, Tang, and Yang}{Lin et~al\mbox{.}}{2021}]%
        {m6}
\bibfield{author}{\bibinfo{person}{Junyang Lin}, \bibinfo{person}{Rui Men},
  \bibinfo{person}{An Yang}, \bibinfo{person}{Chang Zhou},
  \bibinfo{person}{Ming Ding}, \bibinfo{person}{Yichang Zhang},
  \bibinfo{person}{Peng Wang}, \bibinfo{person}{Ang Wang}, \bibinfo{person}{Le
  Jiang}, \bibinfo{person}{Xianyan Jia}, \bibinfo{person}{Jie Zhang},
  \bibinfo{person}{Jianwei Zhang}, \bibinfo{person}{Xu Zou},
  \bibinfo{person}{Zhikang Li}, \bibinfo{person}{Xiaodong Deng},
  \bibinfo{person}{Jie Liu}, \bibinfo{person}{Jinbao Xue},
  \bibinfo{person}{Huiling Zhou}, \bibinfo{person}{Jianxin Ma},
  \bibinfo{person}{Jin Yu}, \bibinfo{person}{Yong Li}, \bibinfo{person}{Wei
  Lin}, \bibinfo{person}{Jingren Zhou}, \bibinfo{person}{Jie Tang}, {and}
  \bibinfo{person}{Hongxia Yang}.} \bibinfo{year}{2021}\natexlab{}.
\newblock \bibinfo{title}{M6: A Chinese Multimodal Pretrainer}.
\newblock
\newblock
\showeprint[arxiv]{2103.00823}


\bibitem[\protect\citeauthoryear{Lin, Maire, Belongie, Bourdev, Girshick, Hays,
  Perona, Ramanan, Zitnick, and Dollár}{Lin et~al\mbox{.}}{2014}]%
        {coco}
\bibfield{author}{\bibinfo{person}{Tsung-Yi Lin}, \bibinfo{person}{Michael
  Maire}, \bibinfo{person}{Serge Belongie}, \bibinfo{person}{Lubomir Bourdev},
  \bibinfo{person}{Ross Girshick}, \bibinfo{person}{James Hays},
  \bibinfo{person}{Pietro Perona}, \bibinfo{person}{Deva Ramanan},
  \bibinfo{person}{C.~Lawrence Zitnick}, {and} \bibinfo{person}{Piotr
  Dollár}.} \bibinfo{year}{2014}\natexlab{}.
\newblock \bibinfo{title}{Microsoft COCO: Common Objects in Context}.
\newblock
\newblock
\showeprint[arxiv]{1405.0312}


\bibitem[\protect\citeauthoryear{Liu, Ott, Goyal, Du, Joshi, Chen, Levy, Lewis,
  Zettlemoyer, and Stoyanov}{Liu et~al\mbox{.}}{2019}]%
        {roberta}
\bibfield{author}{\bibinfo{person}{Yinhan Liu}, \bibinfo{person}{Myle Ott},
  \bibinfo{person}{Naman Goyal}, \bibinfo{person}{Jingfei Du},
  \bibinfo{person}{Mandar Joshi}, \bibinfo{person}{Danqi Chen},
  \bibinfo{person}{Omer Levy}, \bibinfo{person}{Mike Lewis},
  \bibinfo{person}{Luke Zettlemoyer}, {and} \bibinfo{person}{Veselin
  Stoyanov}.} \bibinfo{year}{2019}\natexlab{}.
\newblock \bibinfo{title}{RoBERTa: A Robustly Optimized BERT Pretraining
  Approach}.
\newblock
\newblock
\showeprint[arxiv]{1907.11692}


\bibitem[\protect\citeauthoryear{Lu, Batra, Parikh, and Lee}{Lu
  et~al\mbox{.}}{2019}]%
        {vilbert}
\bibfield{author}{\bibinfo{person}{Jiasen Lu}, \bibinfo{person}{Dhruv Batra},
  \bibinfo{person}{Devi Parikh}, {and} \bibinfo{person}{Stefan Lee}.}
  \bibinfo{year}{2019}\natexlab{}.
\newblock \showarticletitle{ViLBERT: Pretraining Task-Agnostic Visiolinguistic
  Representations for Vision-and-Language Tasks}. In
  \bibinfo{booktitle}{\emph{NIPS}}.
\newblock


\bibitem[\protect\citeauthoryear{Luo, Ji, Shi, Huang, Duan, Li, Li, Bharti, and
  Zhou}{Luo et~al\mbox{.}}{2020}]%
        {univl}
\bibfield{author}{\bibinfo{person}{Huaishao Luo}, \bibinfo{person}{Lei Ji},
  \bibinfo{person}{Botian Shi}, \bibinfo{person}{Haoyang Huang},
  \bibinfo{person}{Nan Duan}, \bibinfo{person}{Tianrui Li},
  \bibinfo{person}{Jason Li}, \bibinfo{person}{Taroon Bharti}, {and}
  \bibinfo{person}{Ming Zhou}.} \bibinfo{year}{2020}\natexlab{}.
\newblock \bibinfo{title}{UniVL: A Unified Video and Language Pre-Training
  Model for Multimodal Understanding and Generation}.
\newblock
\newblock
\showeprint[arxiv]{1906.05743}


\bibitem[\protect\citeauthoryear{Miech, Zhukov, Alayrac, Tapaswi, Laptev, and
  Sivic}{Miech et~al\mbox{.}}{2019}]%
        {howto100}
\bibfield{author}{\bibinfo{person}{Antoine Miech}, \bibinfo{person}{Dimitri
  Zhukov}, \bibinfo{person}{Jean-Baptiste Alayrac}, \bibinfo{person}{Makarand
  Tapaswi}, \bibinfo{person}{Ivan Laptev}, {and} \bibinfo{person}{Josef
  Sivic}.} \bibinfo{year}{2019}\natexlab{}.
\newblock \showarticletitle{HowTo100M: Learning a Text-Video Embedding by
  Watching Hundred Million Narrated Video Clips}. In
  \bibinfo{booktitle}{\emph{ICCV}}. IEEE.
\newblock


\bibitem[\protect\citeauthoryear{Pan, Li, Luo, Xu, Yao, and Mei}{Pan
  et~al\mbox{.}}{2020}]%
        {autogif}
\bibfield{author}{\bibinfo{person}{Yingwei Pan}, \bibinfo{person}{Yehao Li},
  \bibinfo{person}{Jianjie Luo}, \bibinfo{person}{Jun Xu},
  \bibinfo{person}{Ting Yao}, {and} \bibinfo{person}{Tao Mei}.}
  \bibinfo{year}{2020}\natexlab{}.
\newblock \bibinfo{title}{Auto-captions on GIF: A Large-scale Video-sentence
  Dataset for Vision-language Pre-training}.
\newblock
\newblock
\showeprint[arxiv]{2007.02375}


\bibitem[\protect\citeauthoryear{Radford, Kim, Hallacy, Ramesh, Goh, Agarwal,
  Sastry, Askell, Mishkin, Clark, Krueger, and Sutskever}{Radford
  et~al\mbox{.}}{2021}]%
        {clip}
\bibfield{author}{\bibinfo{person}{Alec Radford}, \bibinfo{person}{Jong~Wook
  Kim}, \bibinfo{person}{Chris Hallacy}, \bibinfo{person}{Aditya Ramesh},
  \bibinfo{person}{Gabriel Goh}, \bibinfo{person}{Sandhini Agarwal},
  \bibinfo{person}{Girish Sastry}, \bibinfo{person}{Amanda Askell},
  \bibinfo{person}{Pamela Mishkin}, \bibinfo{person}{Jack Clark},
  \bibinfo{person}{Gretchen Krueger}, {and} \bibinfo{person}{Ilya Sutskever}.}
  \bibinfo{year}{2021}\natexlab{}.
\newblock \bibinfo{title}{Learning Transferable Visual Models From Natural
  Language Supervision}.
\newblock
\newblock
\showeprint[arxiv]{2103.00020}


\bibitem[\protect\citeauthoryear{Raffel, Shazeer, Roberts, Lee, Narang, Matena,
  Zhou, Li, and Liu}{Raffel et~al\mbox{.}}{2019}]%
        {t5}
\bibfield{author}{\bibinfo{person}{Colin Raffel}, \bibinfo{person}{Noam
  Shazeer}, \bibinfo{person}{Adam Roberts}, \bibinfo{person}{Katherine Lee},
  \bibinfo{person}{Sharan Narang}, \bibinfo{person}{Michael Matena},
  \bibinfo{person}{Yanqi Zhou}, \bibinfo{person}{Wei Li}, {and}
  \bibinfo{person}{Peter~J. Liu}.} \bibinfo{year}{2019}\natexlab{}.
\newblock \bibinfo{title}{Exploring the Limits of Transfer Learning with a
  Unified Text-to-Text Transformer}.
\newblock
\newblock
\showeprint[arxiv]{1910.10683}


\bibitem[\protect\citeauthoryear{Ramesh, Pavlov, Goh, Gray, Voss, Radford,
  Chen, and Sutskever}{Ramesh et~al\mbox{.}}{2021}]%
        {dalle}
\bibfield{author}{\bibinfo{person}{Aditya Ramesh}, \bibinfo{person}{Mikhail
  Pavlov}, \bibinfo{person}{Gabriel Goh}, \bibinfo{person}{Scott Gray},
  \bibinfo{person}{Chelsea Voss}, \bibinfo{person}{Alec Radford},
  \bibinfo{person}{Mark Chen}, {and} \bibinfo{person}{Ilya Sutskever}.}
  \bibinfo{year}{2021}\natexlab{}.
\newblock \bibinfo{title}{Zero-Shot Text-to-Image Generation}.
\newblock
\newblock
\showeprint[arxiv]{2102.12092}


\bibitem[\protect\citeauthoryear{Rohrbach, Rohrbach, Tandon, and
  Schiele}{Rohrbach et~al\mbox{.}}{2015}]%
        {mpii}
\bibfield{author}{\bibinfo{person}{Anna Rohrbach}, \bibinfo{person}{Marcus
  Rohrbach}, \bibinfo{person}{Niket Tandon}, {and} \bibinfo{person}{Bernt
  Schiele}.} \bibinfo{year}{2015}\natexlab{}.
\newblock \bibinfo{title}{A Dataset for Movie Description}.
\newblock
\newblock
\showeprint[arxiv]{1501.02530}


\bibitem[\protect\citeauthoryear{Su, Zhu, Cao, Li, Lu, Wei, and Dai}{Su
  et~al\mbox{.}}{2020}]%
        {vlbert}
\bibfield{author}{\bibinfo{person}{Weijie Su}, \bibinfo{person}{Xizhou Zhu},
  \bibinfo{person}{Yue Cao}, \bibinfo{person}{Bin Li}, \bibinfo{person}{Lewei
  Lu}, \bibinfo{person}{Furu Wei}, {and} \bibinfo{person}{Jifeng Dai}.}
  \bibinfo{year}{2020}\natexlab{}.
\newblock \showarticletitle{VL-BERT: Pre-training of Generic Visual-Linguistic
  Representations}. In \bibinfo{booktitle}{\emph{ICLR}}.
\newblock


\bibitem[\protect\citeauthoryear{Sun, Baradel, Murphy, and Schmid}{Sun
  et~al\mbox{.}}{2019a}]%
        {cbt}
\bibfield{author}{\bibinfo{person}{Chen Sun}, \bibinfo{person}{Fabien Baradel},
  \bibinfo{person}{Kevin Murphy}, {and} \bibinfo{person}{Cordelia Schmid}.}
  \bibinfo{year}{2019}\natexlab{a}.
\newblock \bibinfo{title}{Learning Video Representations using Contrastive
  Bidirectional Transformer}.
\newblock
\newblock
\showeprint[arxiv]{2002.06353}


\bibitem[\protect\citeauthoryear{Sun, Myers, Vondrick, Murphy, and Schmid}{Sun
  et~al\mbox{.}}{2019b}]%
        {videobert}
\bibfield{author}{\bibinfo{person}{Chen Sun}, \bibinfo{person}{Austin Myers},
  \bibinfo{person}{Carl Vondrick}, \bibinfo{person}{Kevin Murphy}, {and}
  \bibinfo{person}{Cordelia Schmid}.} \bibinfo{year}{2019}\natexlab{b}.
\newblock \showarticletitle{Videobert: A joint model for video and language
  representation learning}. In \bibinfo{booktitle}{\emph{ICCV}}. IEEE.
\newblock


\bibitem[\protect\citeauthoryear{Sun, Wang, Li, Feng, Tian, Wu, and Wang}{Sun
  et~al\mbox{.}}{2020}]%
        {ERNIE}
\bibfield{author}{\bibinfo{person}{Yu Sun}, \bibinfo{person}{Shuohuan Wang},
  \bibinfo{person}{Yukun Li}, \bibinfo{person}{Shikun Feng},
  \bibinfo{person}{Hao Tian}, \bibinfo{person}{Hua Wu}, {and}
  \bibinfo{person}{Haifeng Wang}.} \bibinfo{year}{2020}\natexlab{}.
\newblock \showarticletitle{ERNIE 2.0: A Continual Pre-Training Framework for
  Language Understanding}. In \bibinfo{booktitle}{\emph{AAAI}}.
\newblock


\bibitem[\protect\citeauthoryear{Szegedy, Ioffe, Vanhoucke, and Alemi}{Szegedy
  et~al\mbox{.}}{2016}]%
        {inceptionv4}
\bibfield{author}{\bibinfo{person}{Christian Szegedy}, \bibinfo{person}{Sergey
  Ioffe}, \bibinfo{person}{Vincent Vanhoucke}, {and} \bibinfo{person}{Alex
  Alemi}.} \bibinfo{year}{2016}\natexlab{}.
\newblock \bibinfo{title}{Inception-v4, Inception-ResNet and the Impact of
  Residual Connections on Learning}.
\newblock
\newblock
\showeprint[arxiv]{1602.07261}


\bibitem[\protect\citeauthoryear{Tan and Bansal}{Tan and Bansal}{2019}]%
        {LXMERT}
\bibfield{author}{\bibinfo{person}{Hao Tan} {and} \bibinfo{person}{Mohit
  Bansal}.} \bibinfo{year}{2019}\natexlab{}.
\newblock \showarticletitle{LXMERT: Learning Cross-Modality Encoder
  Representations from Transformers}. In \bibinfo{booktitle}{\emph{EMNLP}}.
\newblock


\bibitem[\protect\citeauthoryear{van~den Oord, Li, and Vinyals}{van~den Oord
  et~al\mbox{.}}{2018}]%
        {infonce}
\bibfield{author}{\bibinfo{person}{Aaron van~den Oord}, \bibinfo{person}{Yazhe
  Li}, {and} \bibinfo{person}{Oriol Vinyals}.} \bibinfo{year}{2018}\natexlab{}.
\newblock \bibinfo{title}{Representation Learning with Contrastive Predictive
  Coding}.
\newblock
\newblock
\showeprint[arxiv]{1807.03748}


\bibitem[\protect\citeauthoryear{Vaswani, Shazeer, Parmar, Uszkoreit, Jones,
  Gomez, Kaiser, and Polosukhin}{Vaswani et~al\mbox{.}}{2017}]%
        {attallyouneed}
\bibfield{author}{\bibinfo{person}{Ashish Vaswani}, \bibinfo{person}{Noam
  Shazeer}, \bibinfo{person}{Niki Parmar}, \bibinfo{person}{Jakob Uszkoreit},
  \bibinfo{person}{Llion Jones}, \bibinfo{person}{Aidan~N. Gomez},
  \bibinfo{person}{Lukasz Kaiser}, {and} \bibinfo{person}{Illia Polosukhin}.}
  \bibinfo{year}{2017}\natexlab{}.
\newblock \showarticletitle{Attention is All You Need}. In
  \bibinfo{booktitle}{\emph{NIPS}}. ACM, \bibinfo{pages}{6000--6010}.
\newblock


\bibitem[\protect\citeauthoryear{Wu, Schuster, Chen, Le, Norouzi, Macherey,
  Krikun, Cao, Gao, Macherey, et~al\mbox{.}}{Wu et~al\mbox{.}}{2016}]%
        {wordpiece}
\bibfield{author}{\bibinfo{person}{Yonghui Wu}, \bibinfo{person}{Mike
  Schuster}, \bibinfo{person}{Zhifeng Chen}, \bibinfo{person}{Quoc~V Le},
  \bibinfo{person}{Mohammad Norouzi}, \bibinfo{person}{Wolfgang Macherey},
  \bibinfo{person}{Maxim Krikun}, \bibinfo{person}{Yuan Cao},
  \bibinfo{person}{Qin Gao}, \bibinfo{person}{Klaus Macherey}, {et~al\mbox{.}}}
  \bibinfo{year}{2016}\natexlab{}.
\newblock \showarticletitle{Google's neural machine translation system:
  Bridging the gap between human and machine translation}.
\newblock \bibinfo{journal}{\emph{arXiv preprint arXiv:1609.08144}}
  (\bibinfo{year}{2016}).
\newblock


\bibitem[\protect\citeauthoryear{Xu, Mei, Yao, and Rui}{Xu
  et~al\mbox{.}}{2016}]%
        {msrvtt}
\bibfield{author}{\bibinfo{person}{Jun Xu}, \bibinfo{person}{Tao Mei},
  \bibinfo{person}{Ting Yao}, {and} \bibinfo{person}{Yong Rui}.}
  \bibinfo{year}{2016}\natexlab{}.
\newblock \showarticletitle{MSR-VTT: A Large Video Description Dataset for
  Bridging Video and Language}. In \bibinfo{booktitle}{\emph{CVPR}}. IEEE.
\newblock


\bibitem[\protect\citeauthoryear{Yang, Dai, Yang, Carbonell, Salakhutdinov, and
  Le}{Yang et~al\mbox{.}}{2019}]%
        {xlnet}
\bibfield{author}{\bibinfo{person}{Zhilin Yang}, \bibinfo{person}{Zihang Dai},
  \bibinfo{person}{Yiming Yang}, \bibinfo{person}{Jaime Carbonell},
  \bibinfo{person}{Ruslan Salakhutdinov}, {and} \bibinfo{person}{Quoc~V. Le}.}
  \bibinfo{year}{2019}\natexlab{}.
\newblock \showarticletitle{XLNet: Generalized Autoregressive Pretraining for
  Language Understanding}. In \bibinfo{booktitle}{\emph{NIPS}}.
\newblock


\bibitem[\protect\citeauthoryear{Zhenzhong~Lan}{Zhenzhong~Lan}{2019}]%
        {albert}
\bibfield{author}{\bibinfo{person}{Sebastian Goodman Kevin Gimpel Piyush Sharma
  Radu~Soricut Zhenzhong~Lan, Mingda~Chen}.} \bibinfo{year}{2019}\natexlab{}.
\newblock \bibinfo{title}{ALBERT: A Lite BERT for Self-supervised Learning of
  Language Representations}.
\newblock
\newblock
\showeprint[arxiv]{1909.11942}


\bibitem[\protect\citeauthoryear{Zhou, Mou, Fan, Pi, Bian, Zhou, Zhu, and
  Gai}{Zhou et~al\mbox{.}}{2018}]%
        {din}
\bibfield{author}{\bibinfo{person}{Guorui Zhou}, \bibinfo{person}{Na Mou},
  \bibinfo{person}{Ying Fan}, \bibinfo{person}{Qi Pi}, \bibinfo{person}{Weijie
  Bian}, \bibinfo{person}{Chang Zhou}, \bibinfo{person}{Xiaoqiang Zhu}, {and}
  \bibinfo{person}{Kun Gai}.} \bibinfo{year}{2018}\natexlab{}.
\newblock \showarticletitle{Deep Interest Evolution Network for Click-Through
  Rate Prediction}. In \bibinfo{booktitle}{\emph{KDD}}. ACM.
\newblock


\bibitem[\protect\citeauthoryear{Zhou, Liu, Cheng, Gan, and Zhang}{Zhou
  et~al\mbox{.}}{2021}]%
        {CUPID}
\bibfield{author}{\bibinfo{person}{Luowei Zhou}, \bibinfo{person}{Jingjing
  Liu}, \bibinfo{person}{Yu Cheng}, \bibinfo{person}{Zhe Gan}, {and}
  \bibinfo{person}{Lei Zhang}.} \bibinfo{year}{2021}\natexlab{}.
\newblock \bibinfo{title}{CUPID: Adaptive Curation of Pre-training Data for
  Video-and-Language Representation Learning}.
\newblock
\newblock
\showeprint[arxiv]{2104.00285}


\end{thebibliography}


\appendix

\section{\textsc{Alivol}-10M}
\begin{figure}
	\centering
	\includegraphics[width=0.48\textwidth]{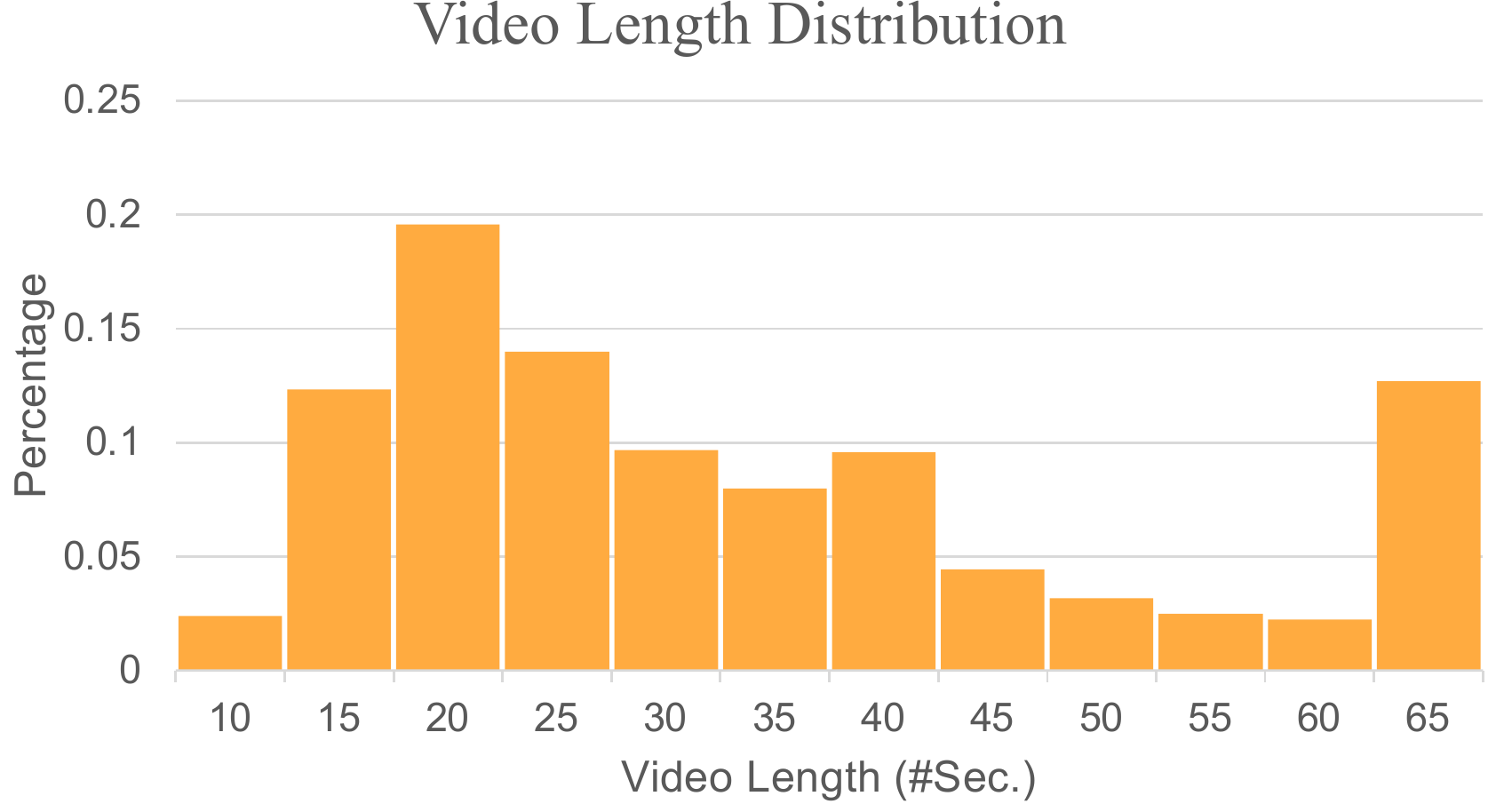}
	\caption{Distributions of video length of \textsc{Alivol}-10M.}
	\label{fig_video_len}
\end{figure}

\begin{figure}
	\centering
	\includegraphics[width=0.48\textwidth]{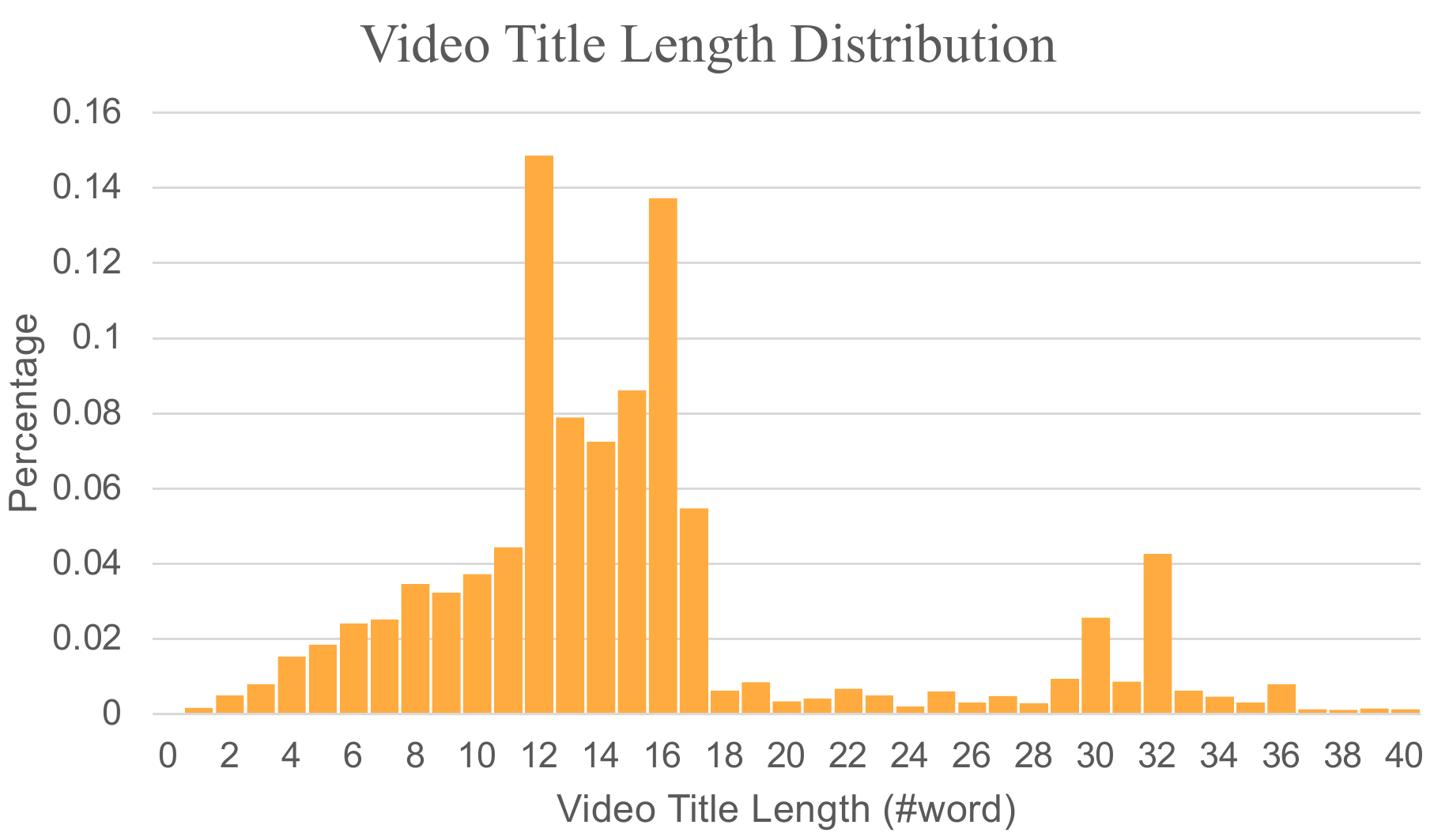}
	\caption{Distributions of video title length of \textsc{Alivol}-10M.}
	\label{fig_title_len}
\end{figure}

In this section, we introduce the distributions of video length and video title length in \textsc{Alivol}-10M dataset. In particular, Figure~\ref{fig_video_len} shows the distribution of video length in \textsc{Alivol}-10M. We can observe that the lengths of most videos are 15-40 seconds, which are the most popular micro-video length at present on online micro-video platforms (\eg TikTok and Kuaishou). Figure~\ref{fig_title_len} shows the distribution of video title length in \textsc{Alivol}-10M. 

\begin{figure*}
	\centering
	\includegraphics[width=0.85\textwidth]{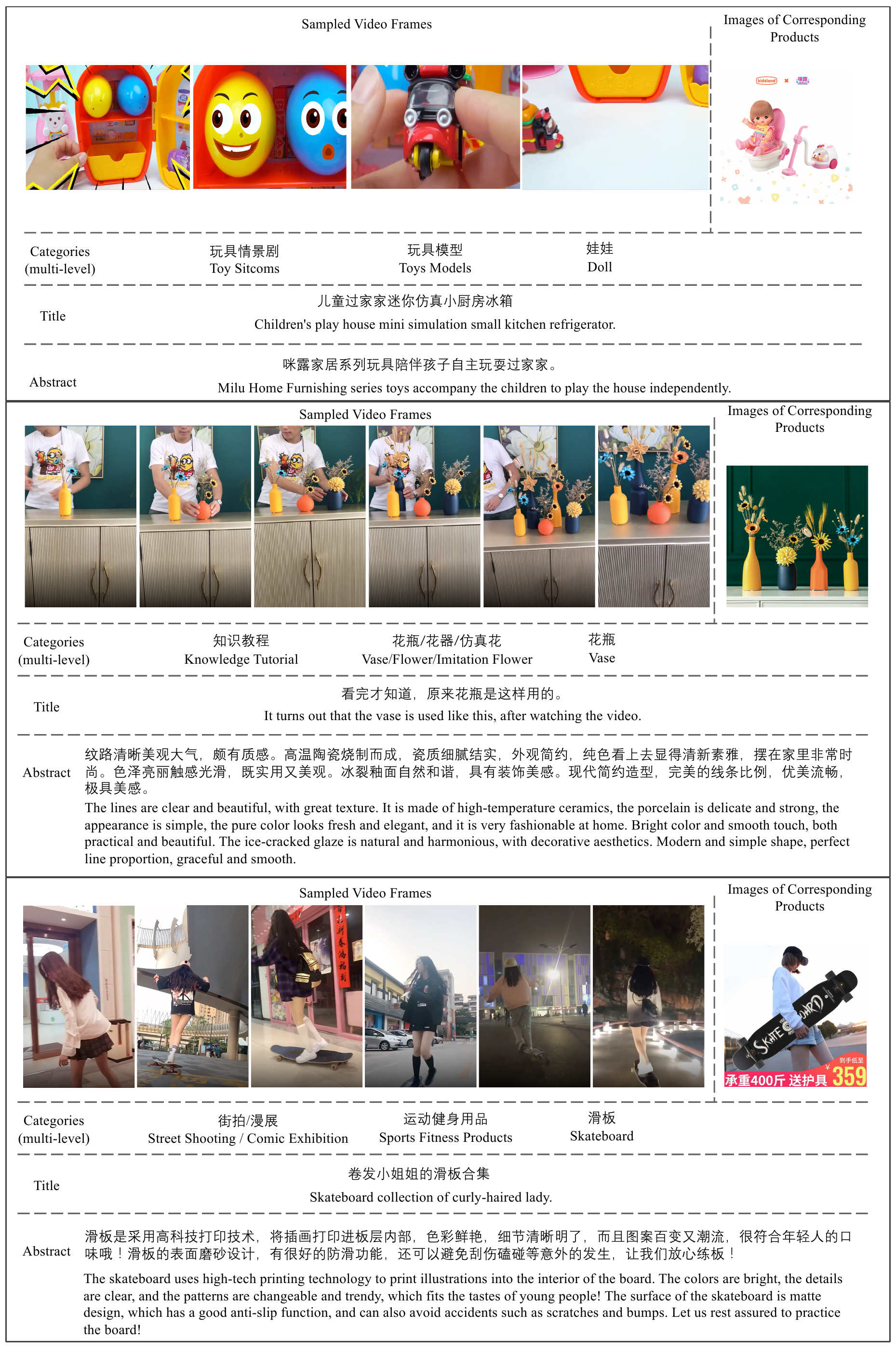}
	\caption{Three example videos of \textsc{Alivol}-10M.}
	\label{datacase}
\end{figure*}

In order to attract users to watch videos, the titles of videos are short and attractive. Figure~\ref{datacase} shows three example videos of \textsc{Alivol}-10M. Besides the video frames and titles, each video also contains three different categories, a corresponding e-commerce product with images, and a long-text abstract. From these three videos, we can observe the following properties of \textsc{Alivol}-10M: 1) high-resolution videos; 2) attractive text including short-text title and long-text abstract, related to video semantically; 3) multi-level categories; 4) associating related products so that more application developments can be carried out.

\section{Case Studies for Cross-Modal Video Retrieval}
\begin{figure*}
	\centering
	\includegraphics[width=0.8\textwidth]{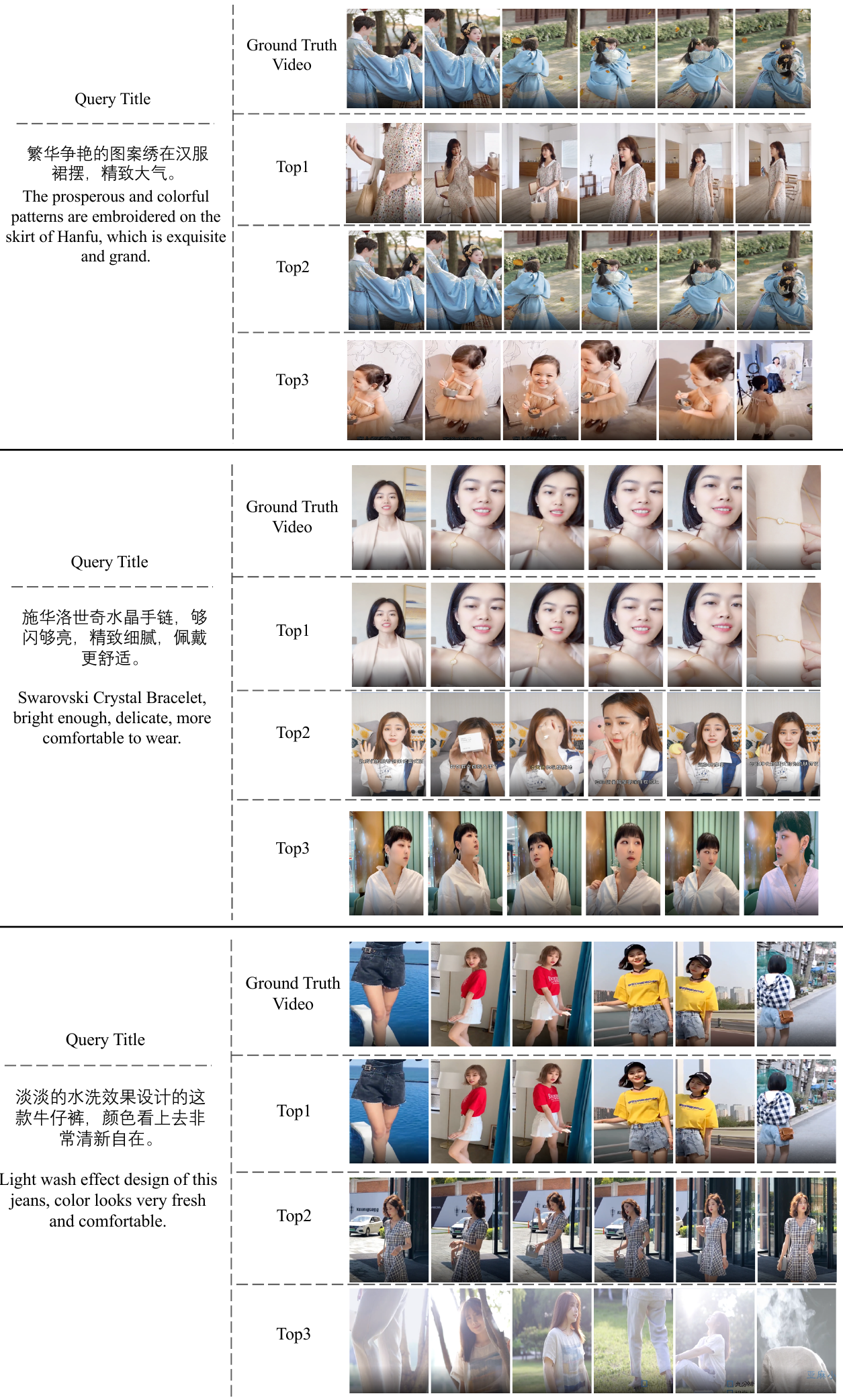}
	\caption{Three example results for the text-based video retrieval task, which is fine-tuned based on our pre-trained \textsc{Victor}.}
	\label{t2v}
\end{figure*}

Figure~\ref{t2v} shows three examples of the text-based video retrieval task based on the pre-trained \textsc{Victor} model. We can note that the pre-trained \textsc{Victor} model recall the ground truth in top 3 results in all three cases. 
Moreover, in the first two cases, we can observe that the other two videos in top-3 recalled results are also relevant to query titles. In the third case, the second result is a bad case, whose product is not jeans. 

\begin{figure*}
	\centering
	\includegraphics[width=0.7\textwidth]{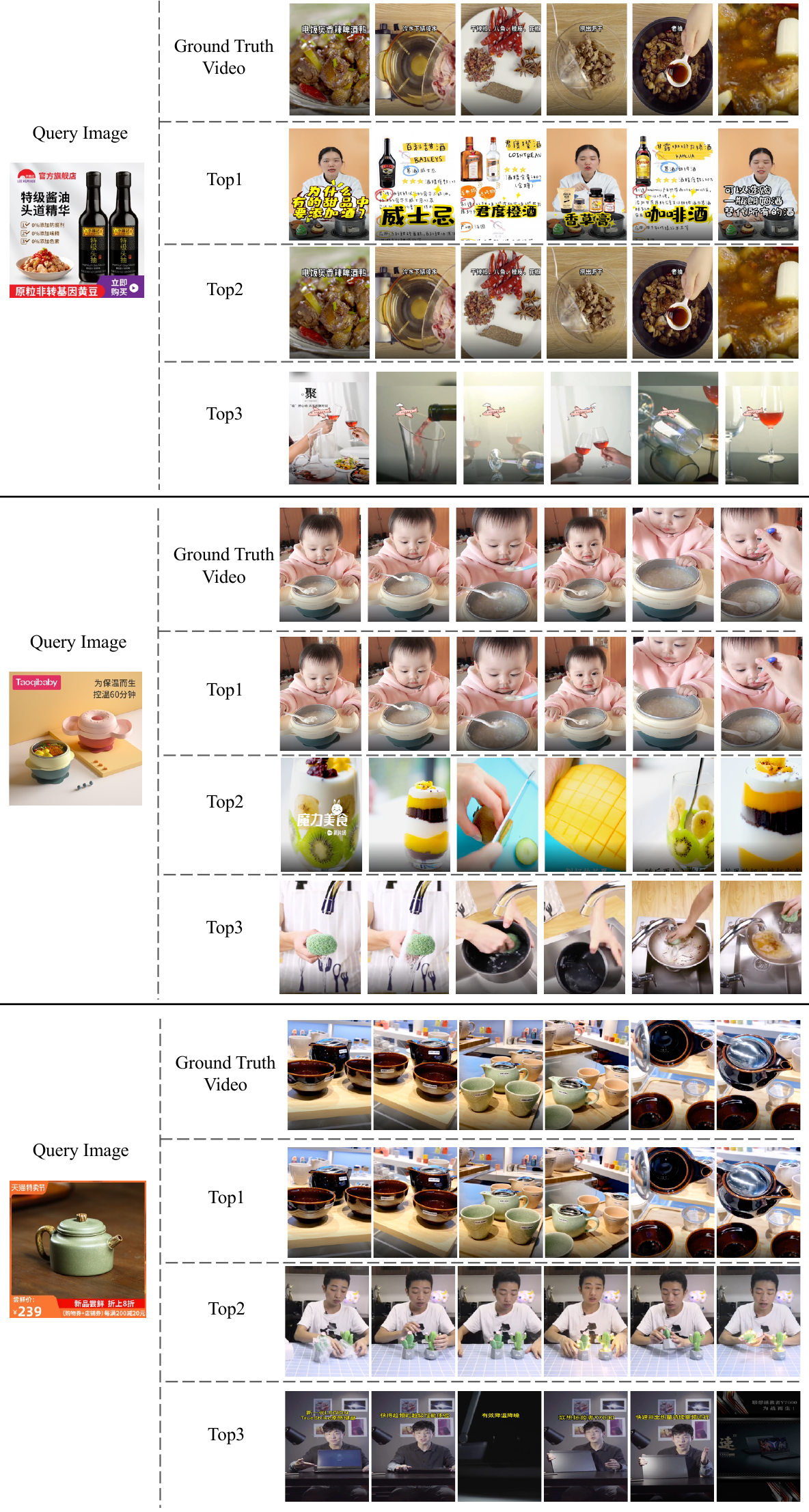}
	\caption{Three example results for the image-based video retrieval task, which is fine-tuned based on our pre-trained \textsc{Victor}.}
	\label{i2v}
\end{figure*}

Figure~\ref{i2v} shows three examples of the image-based video retrieval task based on the pre-trained \textsc{Victor} model. In the first two cases, all the recalled videos are relevant to the query image. However, in the third case, the third retrieved video describes a laptop instead of a teapot. Over all, Figure~\ref{t2v} and Figure~\ref{i2v} demonstrate the capacity of the pre-trained \textsc{Victor} model in the retrieval tasks.

\section{Case Studies for Multimodal Video Caption}
\begin{figure*}
	\centering
	\includegraphics[width=0.85\textwidth]{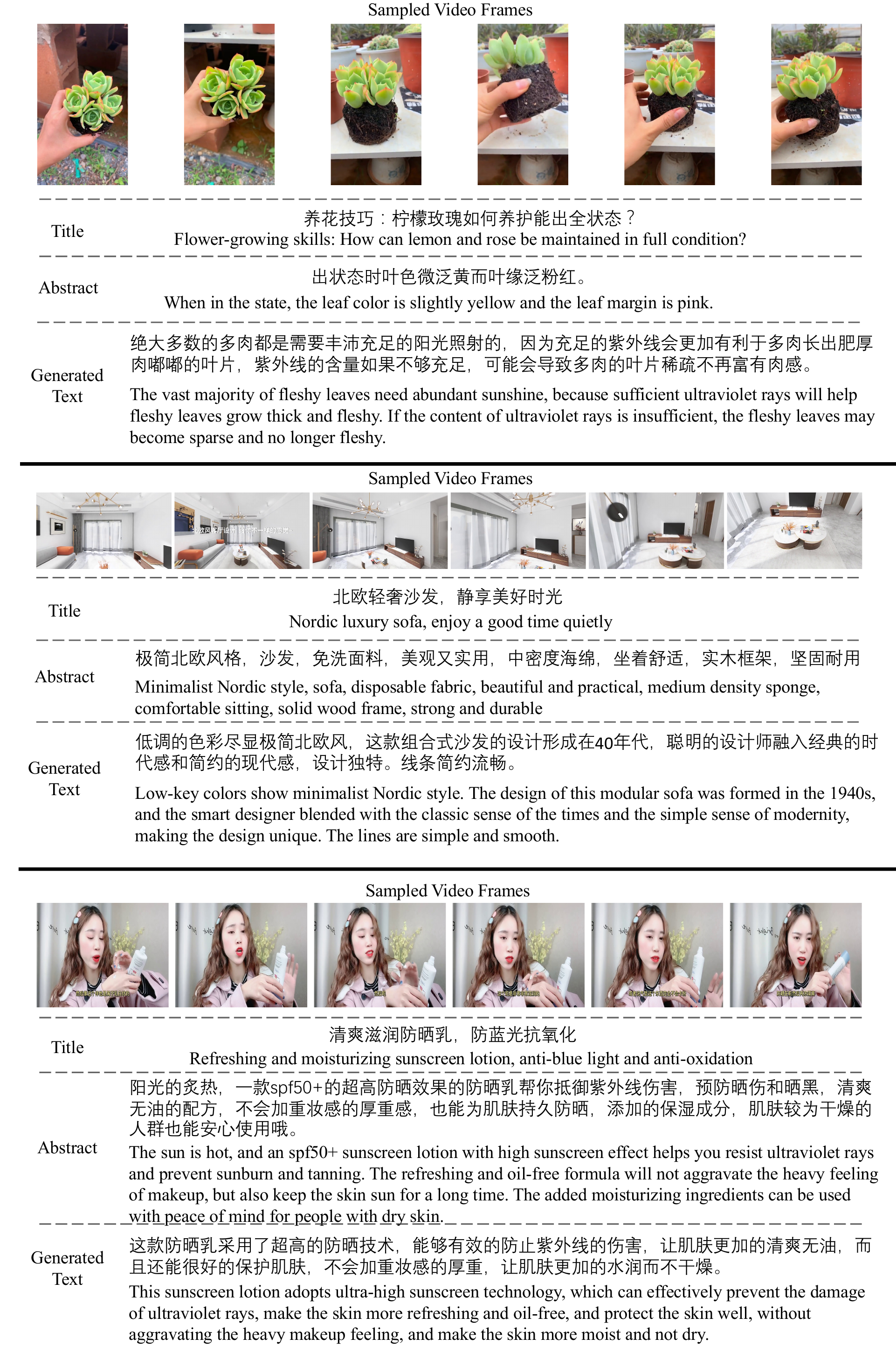}
	\caption{Three example results for the multimodal video caption task, which is fine-tuned based on our pre-trained \textsc{Victor}.}
	\label{cap}
\end{figure*}

Figure~\ref{cap} shows three examples of multimodal video caption task based on the pre-trained \textsc{Victor} model. From the Figure~\ref{cap}, we can note that the generated text is semantically related to the video content, with correct grammar and fluent language. Moreover, it also has the same attractive language style as video abstracts.

\end{document}